\documentclass[runningheads]{llncs}

\usepackage{amsmath,amssymb} 
\usepackage{graphicx}
\usepackage{tikz}
\usepackage{comment} 
\usepackage{color}
\usepackage{bbm}
\usepackage{booktabs}
\usepackage{array}
\usepackage{makecell}
\usepackage{microtype}
\usepackage{siunitx}
\usepackage{cite}
\usepackage{xspace}
\usepackage{hyperref}

\usepackage{xcolor}
\hypersetup{
    colorlinks,
    linkcolor={red!50!black},
    citecolor={blue!50!black},
    urlcolor={blue!80!black}
}

\newcolumntype{P}[1]{>{\centering\arraybackslash}p{#1}}
\newcommand{\PAR}[1]{\vskip4pt \noindent{\bf #1~}}

\newcommand{\grovetwo}{Grove v2}
\newcommand{\nameDataset}{RIO10}
\newcommand{\nameEvalMetricLong}{Dense Correspondence Re-Projection Error}
\newcommand{\nameEvalMetricShort}{DCRE}

\makeatletter
\DeclareRobustCommand\onedot{\futurelet\@let@token\@onedot}
\def\@onedot{\ifx\@let@token.\else.\null\fi\xspace}
\def\eg{\emph{e.g}\onedot} 
\def\ie{\emph{i.e}\onedot} 
\def\cf{\emph{c.f}\onedot}

\def\etal{\emph{et al}\onedot}
\makeatother

\begin{document}
\pagestyle{headings}
\mainmatter
\def\ECCVSubNumber{287} 

\title{Beyond Controlled Environments: 3D Camera Re-Localization in Changing Indoor Scenes}

\titlerunning{3D Camera Re-Localization in Changing Indoor Scenes}

\author{Johanna Wald\inst{1} \and
Torsten Sattler\inst{2,3} \and
Stuart Golodetz\inst{4} \and\\
Tommaso Cavallari\inst{4} \and
Federico Tombari \inst{1,5}}
\authorrunning{Wald et al.}
\institute{Technical University of Munich \and 
Chalmers University of Technology \and 
CIIRC, Czech Technical University in Prague \and
Five AI Ltd. \and Google Inc.}
\maketitle

\begin{abstract}
Long-term camera re-localization is an important task with numerous computer vision and robotics applications. Whilst various outdoor benchmarks exist that target lighting, weather and seasonal changes, far less attention has been paid to appearance changes that occur indoors. This has led to a mismatch between popular indoor benchmarks, which focus on static scenes, and indoor environments that are of interest for many real-world applications. In this paper, we adapt \emph{3RScan} -- a recently introduced indoor RGB-D dataset designed for object instance re-localization -- to create \emph{\nameDataset}, a new long-term camera re-localization benchmark focused on indoor scenes. We propose new metrics for evaluating camera re-localization and explore how state-of-the-art camera re-localizers perform according to these metrics. We also examine in detail how different types of scene change affect the performance of different methods, based on novel ways of detecting such changes in a given RGB-D frame. Our results clearly show that long-term indoor re-localization is an unsolved problem. Our benchmark and tools are publicly available at \url{waldjohannau.github.io/RIO10}.
\end{abstract}

\begin{figure}[!t]
    \centering
    \includegraphics[width=0.94\linewidth]{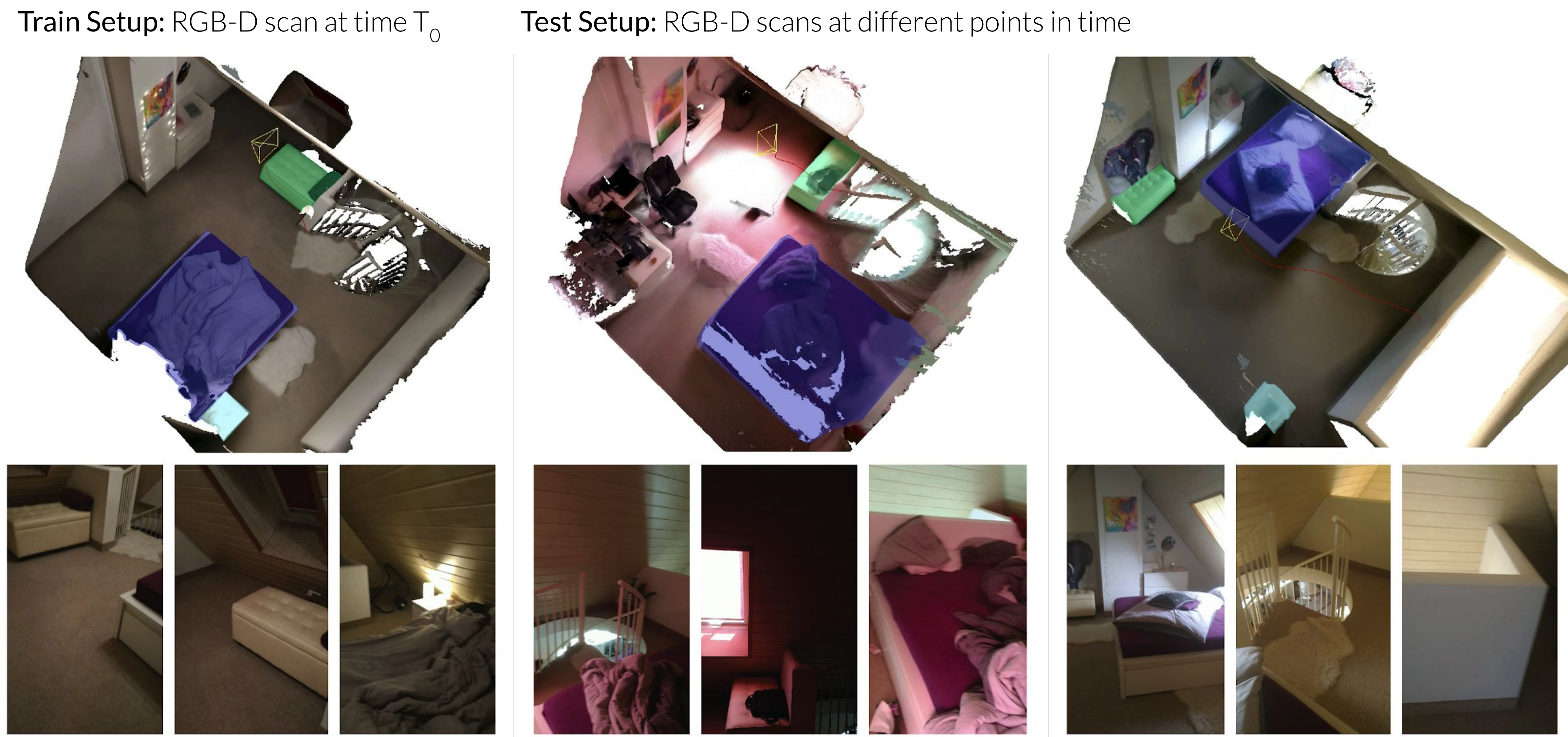} 
    \caption{Visual re-localization in changing indoor scenes: we introduce a new benchmark based on \emph{3RScan} \cite{Wald2019RIO}, together with a new evaluation methodology, for measuring 6DoF re-localization performance given a reference RGB(-D) sequence of an indoor scene at time $T_0$ (left), and query sequences taken at different points in time (center and right).}
    \label{fig:teaser}
\end{figure}

\setcounter{footnote}{0}

\section{Introduction}
Visual re-localization is the problem of estimating the precise position and orientation from which a given image was taken with respect to a known scene. It is a key component of advanced computer vision applications such as AR/VR~\cite{Castle2008,Paucher2010,Golodetz2015SPDEMO,Valentin2015SP,Lynen2015RSS,Rodas2015,Bae2016}, and robotics systems such as self-driving cars~\cite{Bresson2017} and drones~\cite{Lim12CVPR}. Real-world scenes are highly dynamic, exhibiting changes in illumination, appearance and/or geometry. These changes are caused by a variety of factors, including the time of day, the presence of artificial light sources and, most prominently, humans interacting with their environments, \eg{} by redecorating a room, or using furniture and objects in day-to-day life. Like human perception, visual re-localization algorithms should be as robust as possible to such changes to enable long-term operation in the real world. However, the datasets traditionally used for evaluating visual re-localization performance~\cite{Li-ECCV-2008,Li-ECCV-2012worldwide,Chen2011CVPR,Irschara09CVPR,Kendall2015,Shotton2013,Valentin2016,Sattler2012BMVC} either do not contain such changes~\cite{Shotton2013,Valentin2016,Kendall2015} or do not provide a means of quantifying their impact \cite{Li-ECCV-2008,Li-ECCV-2012worldwide,Chen2011CVPR,Irschara09CVPR,Sattler2012BMVC,Taira2018inloc}. 
Only recently released datasets such as Aachen Day-Night~\cite{Sattler2018,Sattler2012BMVC}, (extended) CMU Seasons~\cite{Sattler2018,Badino_IV11}, RobotCar Seasons~\cite{Sattler2018,Maddern2017}, and SILDa~\cite{SILDa} explicitly model such changes. By providing a reference representation and test images taken under different conditions, the corresponding works point out failure cases of existing re-localization algorithms, in turn motivating the community to devise more robust methods~\cite{Dusmanu2019CVPR,Sarlin2019coarse,Toft2018ECCV,Larsson2019ICCV,Germain2019sparsetodense,Toft2017,Anoosheh2019ICRA,Porav2018ICRA,Benbihi2019ICCV}. However, these datasets mostly focus on outdoor scenes, where most changes are cyclic (\eg{} day-night, seasonal and weather changes) and can thus easily be predicted by neural networks~\cite{Porav2018ICRA,Anoosheh2019ICRA}.

As shown in Fig.~\ref{fig:teaser}, indoor scenes are arguably more diverse, and exhibit changes -- including complex illumination changes, as well as geometric and appearance variations caused by human interaction -- that are harder to predict. The only indoor datasets exhibiting these types of changes~\cite{Taira2018inloc,CarlevarisBianco2016} were captured in public spaces, where there is limited human interaction with the environment. Just as importantly, these datasets do not quantify changes, and do not provide the means to measure their impact on re-localization performance. They therefore cannot be used to measure to what degree existing re-localization algorithms are able to handle realistic changes occurring in everyday indoor scenes.

\PAR{}This paper makes the following \textbf{contributions}: 
(\textbf{1}) We construct an \textbf{indoor re-localization benchmark} based on a recently released dataset, \emph{3RScan}~\cite{Wald2019RIO}. 
\emph{3RScan} captures everyday scenes over a long period of time (\cf Fig.~\ref{fig:teaser}) and thus depicts a wide range of changes not captured by other datasets from the literature. 
(\textbf{2}) We propose a \textbf{novel framework to quantify changes in (indoor) scenes}, covering appearance, geometric, and semantic changes. This enables us, for what to the best of our knowledge is the first time, to quantifiably measure the impact of different types of change on the accuracy of the camera poses predicted by visual re-localization algorithms. 
(\textbf{3}) We evaluate state-of-the-art methods for re-localization in static indoor and changing outdoor scenes, and show through detailed experiments that indoor re-localization in real-world scenes is far from being solved. (\textbf{4}) Based on our experiments, we propose a set of \textbf{open challenges} for the community to work on. We make our benchmark, framework, and evaluation protocols \textbf{publicly available}. We think this benchmark closes a gap in the literature by going beyond controlled indoor environments, similar to recent high-impact benchmarks modelling outdoor scene changes~\cite{Sattler2018,Maddern2017,Badino_IV11,Sattler2012BMVC}.

\section{Related Work}

\PAR{Benchmarks.} A variety of datasets exist to target different aspects of the camera re-localization problem (\cf Tab.~\ref{table:relocalization_datasets})\footnote{We exclude other semantic indoor~\cite{Silberman2012,Song2015,Hua2016,Armeni2017,Chang2017,Dai2017} and submap merging~\cite{Golodetz2018} datasets that are neither designed for camera re-localization, nor include scene changes. We also exclude outdoor datasets unsuited to measure re-localization performance in changing scenes~\cite{Kendall2015,Chen2011,Li-ECCV-2008,Li-ECCV-2012worldwide,Irschara09CVPR}, and purely synthetic datasets~\cite{Saeedi2019,Li2018BMVC}.}. 
For the task of re-localizing in \emph{outdoor} scenes that change over time, a multitude of benchmarks exists that look at day vs.\ night changes, season and weather changes, and long-term geometric changes based on e.g.\ changing vegetation or construction projects. \emph{Aachen Day-Night} \cite{Sattler2018} extends the Aachen dataset \cite{Sattler2012BMVC} to support evaluation of a re-localizer's ability to estimate the poses of night-time, outdoor, RGB-only images against a day-time 3D model.
\emph{RobotCar Seasons} \cite{Sattler2018} is based on a subset of the outdoor Oxford RobotCar dataset \cite{Maddern2017}. It focuses on re-localization across different seasons and weather conditions, but also contains a challenge related to localizing low-quality night-time images. While RobotCar Seasons covers an urban region, the (extended) \emph{CMU-Seasons} dataset~\cite{Sattler2018,Badino_IV11} also covers more vegetated outdoor scenes. \emph{SILDa} \cite{SILDa} depicts a small block of buildings in London and provides test images under changing conditions such as weather and illumination changes. 

For many years, the most popular \emph{indoor} datasets have been \emph{7-Scenes} \cite{Shotton2013} and \emph{12-Scenes} \cite{Valentin2016}, which only contain static scenes and exhibit no changes between train and test time. There do exist indoor datasets containing changes, e.g.\ \emph{InLoc} \cite{Taira2018inloc} consists of non-sequential RGB-D training images that are registered to floor plans of university buildings~\cite{Wijmans2017}, and RGB-only query images taken at a later date by hand-held devices. 
Moreover, \emph{InLoc} and \emph{NCLT} \cite{CarlevarisBianco2016} both contain scene changes such as moved objects. However, neither provide any means of quantifying the impact that different changes have on re-localization performance. In this paper, we address this problem by introducing a novel framework to properly quantify the effects of changes in indoor scenes. 

\begin{table}[!t]
	\centering
	\caption{Overview of Camera Re-Localization Benchmarks}
	\label{table:relocalization_datasets}
	\resizebox{\textwidth}{!}
	{
	    \scriptsize{
		{\renewcommand{\arraystretch}{1.3}%
			\begin{tabular}{lcccccc}
				\toprule
				\textbf{Dataset} & \textbf{Train Images } & \textbf{ Test/Val Images } & \textbf{Setup} & \textbf{Sequential} & \textbf{Time Span} \\
				\midrule
				7-Scenes \cite{Glocker2013RealtimeRC}  & 26000 & 17000 & indoor & yes & no\\
				12-Scenes \cite{Valentin2016} & 16926 & 5702 & indoor & yes & no\\
				InLoc \cite{Taira2018inloc} & 9972 & 329 & indoor & no & few days\\
				Aachen Day-Night \cite{Sattler2018} & 4328 & 922 & outdoor & no & few years\\
				Extended CMU-Seasons \cite{Sattler2018} & 60937 & 56613 & outdoor & yes & 2 years\\
				RobotCar Seasons \cite{Sattler2018} & 26121 & 11934 & outdoor & yes & 1 year\\
				SILDa \cite{SILDa} & 8334 & 6064 & outdoor & yes & 1 year\\
				NCLT \cite{CarlevarisBianco2016} & N/A & N/A & both & yes & 15 months\\
				\midrule
				\textbf{\nameDataset{ }(Ours)} &  52562 & 200159 & indoor & yes & 1 year \\
				\bottomrule
	\end{tabular}}}}
\end{table}

\PAR{Camera re-localization methods} can be broadly divided into four types:

\noindent \emph{Image retrieval} methods typically match the query image against images with known poses in a database \cite{GalvezLopez2011,Glocker2015},
but can struggle to generalise to novel poses.
Strategies to mitigate this include the use of synthesized views~\cite{Gee2012,Torii2015}, interpolation between database poses~\cite{Balntas2018,Laskar2017,Torii2011,Zamir2010ECCV}, and triangulation based on relative poses~\cite{Zhang2006TDPVT,Zhou2019ARXIV}.
To achieve scalability in terms of memory and run-time, place recognition methods \cite{Torii2015,Arandjelovic2016}
typically use compact image-level descriptors. Such methods perform well under appearance and limited viewpoint changes~\cite{Sattler2018}. 

\noindent \emph{Direct pose regression} methods, which aim to directly regress a pose from the query image, are often based on pose regression networks \cite{Kendall2015,Kendall2016,Kendall2017,Melekhov2017,Wu2017,Acharya2019}, although decision forest \cite{Kacete2017}, GAN \cite{Bui2019} and LSTM \cite{Clark2017,Walch2017} variants also exist. On the whole, they have not yet matched the precision of state-of-the-art structure-based and RGB-D methods indoors. Recent work by Sattler \etal\ \cite{Sattler2019} has suggested that they are conceptually similar to image retrieval, and may thus face ongoing challenges in generalising to novel poses and achieving highly accurate pose predictions. Some direct pose regression methods \cite{Brahmbhatt2018MapNet,Valada2018,Radwan2018,Li2019arXiv} now exploit the relative poses between images to improve accuracy, and in some cases \cite{Valada2018,Radwan2018} have achieved accuracies that are competitive with state-of-the-art RGB-D methods. However, thus far they have had to rely on estimated poses from previous frames, making them effectively camera tracking approaches that are incomparable with methods that are able to re-localize from only a single image.

\noindent \emph{Structure-based} methods typically match 2D features in the image with 3D points in the scene, and then pass the correspondences to a RANSAC-based backend for camera pose estimation. A classic example is Active Search \cite{Sattler2017}, which performs efficient bidirectional matching using SIFT-based visual vocabularies. Hierarchical localization methods~\cite{Irschara09CVPR,Sattler2012,Sarlin2019coarse,Taira2018inloc,Taira2019} use an initial image retrieval step to make matching more efficient, \ie they first determine a set of potentially visible locations and restrict 2D-3D matching to these. For long-term localization under changing conditions, state-of-the-art methods typically rely on learned  features~\cite{Dusmanu2019CVPR,Germain2019sparsetodense,Sarlin2019coarse,Widya2018CVA}, \eg~HF-Net \cite{Sarlin2019coarse} uses sparse SuperPoint \cite{DeTone2018} and DOAP \cite{He2018} features, whilst \cite{Dusmanu2019CVPR} uses sparse higher-level features extracted from deeper layers of a CNN. 
Both achieve state-of-the-art results on outdoor benchmarks from~\cite{Sattler2018} and outperform approaches based on dense feature matching~\cite{Taira2018inloc,Germain2019sparsetodense,Widya2018CVA}. Another popular approach to outdoor long-term localization is to use semantic information~\cite{Toft2017,Toft2018ECCV,Larsson2019ICCV,Schoenberger2018}. However, \cite{Taira2019} argues that most of these approaches are not directly applicable to indoor scenes. 
Similarly, object-based localization methods~\cite{SalasMoreno2013CVPR,Li2019ICRA,Ardeshir2014ECCV,Atanasov2016IJRR} do not seem applicable in the context of re-localization in changing indoor scenes, as many objects are likely to change their position.

\noindent \emph{Scene coordinate regression (SCoRe)} methods densely regress the scene coordinates of query image pixels using a regression forest \cite{Shotton2013,GuzmanRivera2014,Valentin2015RF,Brachmann2016,Meng2016,Cavallari2017,Meng2017IROS,Golodetz2018,Meng2018IROS,Cavallari2019PAMI,Cavallari20193DV}, a neural network \cite{Brachmann2017,Brachmann2018,Duong2018,Li2018RSS,Li2018ECCV,Brachmann2019neural,Brachmann2019esac,Yang2019,Brachmann2020}, or both \cite{Massiceti2017}. The correspondences are used to generate pose hypotheses using PnP/Kabsch that are then refined using RANSAC. These methods can be categorised based on whether they expect RGB \cite{Brachmann2016,Meng2016,Brachmann2017,Meng2017IROS,Brachmann2018,Duong2018,Li2018RSS,Li2018ECCV,Brachmann2019neural,Brachmann2019esac,Yang2019} or RGB-D \cite{Shotton2013,GuzmanRivera2014,Valentin2015RF,Cavallari2017,Meng2018IROS,Cavallari2019PAMI,Cavallari20193DV} input at test time, and whether they require offline training (most methods) or can be used online \cite{Cavallari2017,Cavallari2019PAMI,Cavallari20193DV}. Better performance has typically been achieved using RGB-D~\cite{Cavallari2019PAMI} rather than RGB-only \cite{Brachmann2018,Brachmann2019esac} input, although RGB-only methods are gradually closing the gap. The state-of-the-art SCoRe relocaliser for indoor RGB-D scenes is currently \grovetwo~\cite{Cavallari2019PAMI}, an online regression forest method, although a network-based variant of this \cite{Cavallari20193DV} performs better outdoors.

A few approaches defy such a categorisation. Valentin \etal\ \cite{Valentin2016} use continuous pose optimisation to refine the results of an initial matching process based on a retrieval forest and multiscale navigation graph. Nakashima \etal\ \cite{Nakashima2019} replace the feature matching step in hierarchical localization with dense regression. Other methods perform retrieval using a point cloud \cite{Deng2016} or 3D model \cite{Lu2016} constructed from multiple query images, or hallucinate a subvolume and match that against a database \cite{Schoenberger2018}. Since our main contribution here is to propose a new benchmark and metrics for evaluating camera re-localization in changing indoor scenes, we focus our attentions on those re-localizers that are known to currently have state-of-the-art performance on static indoor scenes or dynamic outdoor scenes, and explore how their performance is affected when the scenes change.

\section{Benchmark Dataset}
\label{sec:benchmark_dataset}

\begin{table}[!t]
\centering
\caption{Scene statistics and images of the reference/train scan of \emph{\nameDataset}.}
\label{table:selected_scenes}
\resizebox{\textwidth}{!}{
{\renewcommand{\arraystretch}{1.3}
\begin{tabular}{p{5.5cm}cccccccccc}
& \multicolumn{10}{c}{\includegraphics[width=0.55\linewidth]{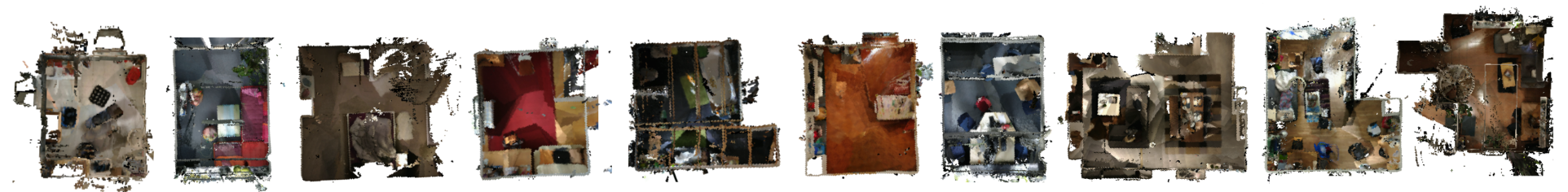}}\\
\toprule
\textbf{Scene} & \textbf{S01} & \textbf{S02} & \textbf{S03} & \textbf{S04} & \textbf{S05} & \textbf{S06} & \textbf{S07} & \textbf{S08} & \textbf{S09} & \textbf{S10} \\
\midrule
Rescans & 6 & 8 & 7 & 10 & 5 & 12 & 8 & 5 & 5 & 8\\
Max Day-Span Between Captures & 176 & 165 & 369 & 176 & 163 & 173 & 104 & 229 & 1 & 168 \\
\# Object Instances & 39 & 33 & 20 & 28 & 44 & 49 & 39 & 61 & 67 & 63 \\ 
\# Changed Object Instances  & 5--9 & 5--6 & 2--3 & 1--5 & 1--2 & 1--6 & 6--10 & 1--5 & 5--6 & 7--9 \\
\bottomrule
\end{tabular}}}
\end{table}

The original \emph{3RScan} \cite{Wald2019RIO}, which was the first large-scale, real-world dataset of changing indoor environments, consists of $1482$ 3D scans of around $450$ natural indoor environments. Each scene has $m$ globally aligned 3D models, each reconstructed from an RGB-D sequence $s$ recorded at time $T_s$, using a hand-held Google Tango phone with camera intrinsics $K_s \in \mathbb{R}^{3 \times 3}$. Reasonably accurate camera poses $\{P_{s,1}, ..., P_{s,k_s}\}$ for each sequence $s$ (of length $k_s$) are determined via an offline bundle adjustment framework, based on fisheye images. A  pose
\begin{equation}
\textstyle P_{s,i} \in \mathbb{R}^{4 \times 4} = \left[\begin{array}{cc}
R_{s,i} & t_{s,i} \\
\mathbf{0}^\top & 1
\end{array}\right]
\end{equation}
is defined by a rotation matrix $R_{s,i} \in \mathbb{R}^{3 \times 3}$ and a translation vector $t_{s,i} \in \mathbb{R}^3$. 
Note that $P_{s,i}$ transforms from the local camera coordinate system to the 3D model coordinate system.  
Whilst originally designed for object re-localization, \emph{3RScan} can also -- when slightly adapted -- enable benchmarking of related tasks such as long-term camera re-localization. Due to the large size of the original dataset, we have chosen to focus on a 10-scene subset of it, which we call \emph{\nameDataset}, for our experiments and evaluation protocol (\cf Tbl.~\ref{table:selected_scenes}). We split the sequences and 3D models into training, validation (one sequence per scene) and testing sets, leaving us with 10 train, 10 validation and 54 test sequences overall. The provided 3D models have both color and semantics (see Fig.\ \ref{fig:rendering}), and are defined as $\{\mathcal{M}_s : 0 \le s < m\}$, where $\mathcal{M}_0$ is our reference/training scan, and each other scan $\mathcal{M}_s$ is a test or validation scan. The 10 scenes chosen for \emph{\nameDataset{}} were selected due partly to their scanning frequency, and partly to their scene and change diversity. Indeed, they are among the scenes in \emph{3RScan} with the highest time span and scanning frequency (5--12 scans each). \emph{\nameDataset{}} features many different indoor scenarios (messy laundry basements, offices or bathrooms) and different types of change (\eg~diversity in lighting, both subtle and significant movements of large/small and rigid/non-rigid objects, and ambiguous changes where objects of the same appearance move). Whilst we decided to evaluate on only a small subset of \emph{3RScan}, the remaining scans are still useful for training future models. To simplify evaluation, each test 3D model and camera pose is provided in the training sequence's reference frame. Due to the low resolution and frame-rate of the raw Tango depth maps, we generated depth renderings of the 3D models for each RGB frame, together with ground-truth 2D instance segmentations. For reproducibility, all data, along with the evaluation tools and per-frame statistics, will be made publicly available.

\section{Evaluating Re-Localization in Changing Indoor Scenes}

Having described our benchmark dataset, we now propose an evaluation methodology for the well-studied camera re-localization problem, as well as novel ways to quantify scene changes. Compared to common evaluation measures from previous camera re-localization benchmarks, we show the advantages of alternative metrics such as the normalised absolute correspondence re-projection error (Sec. \ref{sec:reprojection_error}) when measuring camera re-localization performance. To analyse how re-localization methods are able to generalise to changes in the scene, we propose various measures to quantify the change in each image. This is important, since it gives us an understanding of whether and how different methods are affected by different types of scene change.

\subsection{Quantifying Change in (Indoor) Scenes}
\label{sec:classify_change}

\begin{figure}[!t]
    \centering
    \includegraphics[width=0.97\linewidth]{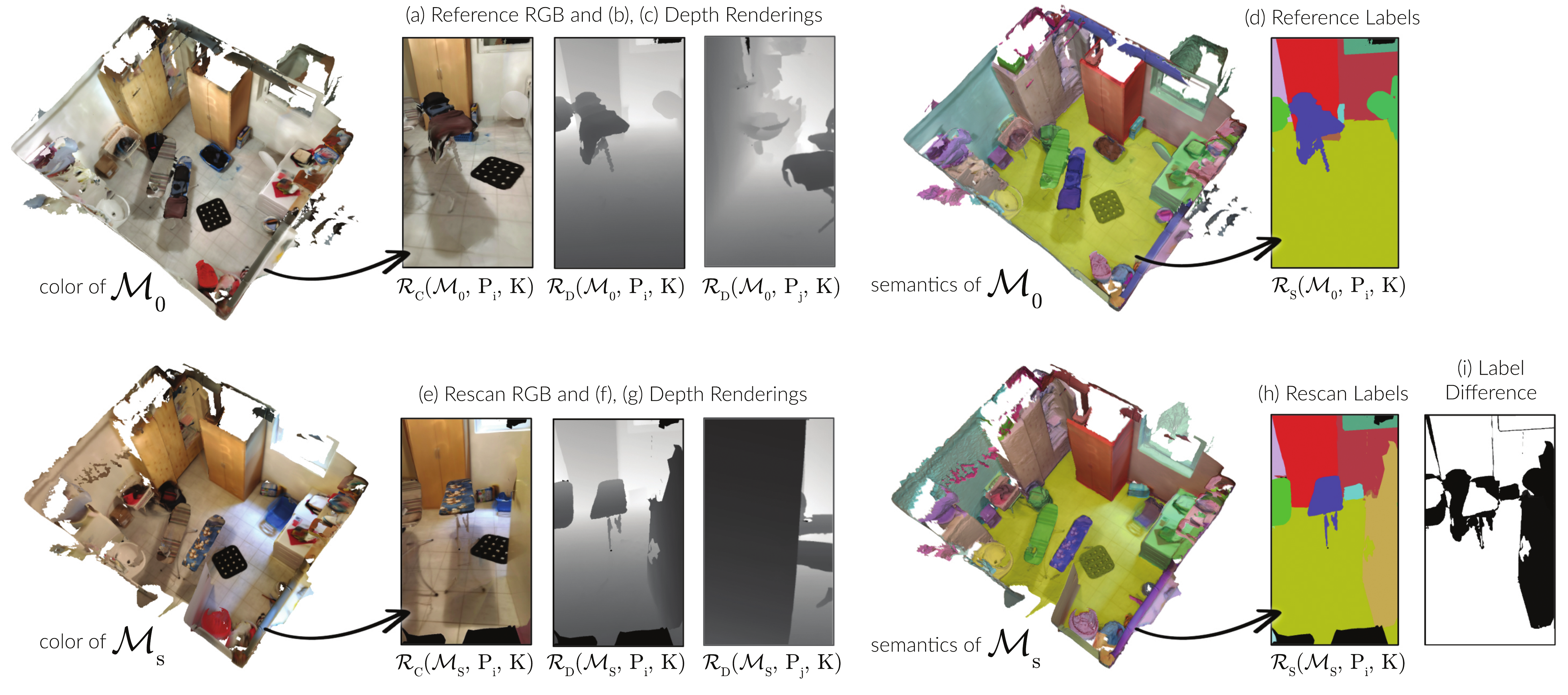}
    \caption{We render synthetic RGB, depth and semantic images from our 3D reference and test models $\mathcal{M}_0$ and $\mathcal{M}_s$, and use them to compute the scene change measures described in Sec.\ \ref{sec:classify_change}. See the main text for details.}
    \label{fig:rendering} 
\end{figure}

In the following, we introduce different measures to quantify the extent to which an RGB-D frame in one of the test sequences has changed with respect to the same view of the reference scan. To compute the measures, we make use of synthetic views of the globally aligned semantic and textured 3D models in \emph{3RScan} (see Sec.\ \ref{sec:benchmark_dataset}). To produce these synthetic views, we define three different rendering functions: $\mathcal{R}_C$ for color, $\mathcal{R}_D$ for depth, and $\mathcal{R}_S$ for semantics. Each of these takes a 3D model $\mathcal{M}$, a pose matrix $P \in \mathbb{R}^{4 \times 4}$, and a camera intrinsics matrix $K \in \mathbb{R}^{3 \times 3}$, and produces a $w \times h$ synthetic view of $\mathcal{M}$ as seen from $P$ using a camera with intrinsics $K$ (see Fig.\ \ref{fig:rendering} for examples).

\subsubsection{Visual Appearance Change}
Given these rendering functions, we can define measures for the visual appearance change between two different models $\mathcal{M}$ and $\mathcal{M'}$ as seen from a given pose $P$ by a camera with intrinsics $K$. Let $I = \mathcal{R}_C(\mathcal{M},P,K)$ and $I' = \mathcal{R}_C(\mathcal{M}',P,K)$ be color renderings of the two models from $P$. Given these, we consider two different measures of the visual appearance change -- the normalized correlation coefficient $\rho_v$, defined as
\begin{equation}
    \label{eq:CCOFF}
    \rho_v = \frac{\sum_u (I(u) - I'(u))^2}{\sqrt{\left( \sum_u I(u)^2 \right) \cdot \left( \sum_u I'(u)^2 \right)}},
\end{equation}
and the normalized sum of squared differences $\zeta_v$, defined as
\begin{equation}
    \label{eq:NSSD}
    \zeta_v =\frac{\sum_u(\bar{I}(u) \cdot \bar{I}'(u))^2}{\sqrt{\sum_u (\bar{I}(u) \cdot \bar{I}'(u))^2}},
\end{equation}
in which $\bar{I}(u) = I(u) - \frac{1}{w\cdot h} \textstyle \sum_{u'}^{} I(u')$. Note that in our experiments, $\mathcal{M}$ is a rescan (see Fig. \ref{fig:rendering}(e)), $\mathcal{M}'$ is the corresponding reference scan (see Fig. \ref{fig:rendering}(a)), and $P$ is a pose from one of the rescan/testing sequences. 

\subsubsection{Semantic Change}
We can also define a semantic change measure $\zeta_s$, based on the percentage of altered pixels in the 2D instance segmentation images. Let $L = \mathcal{R}_S(\mathcal{M},P,K)$ and $L' = \mathcal{R}_S(\mathcal{M}',P,K)$ be semantic renderings of the two models from $P$, and $V^{(s)}_{\{L,L'\}}$ be the set of pixels that have a valid instance ID in both $L$ and $L'$. Then we can define

\begin{equation}
    \textstyle
    \label{eq:semantic_change}
    \zeta_s = \frac{1}{\left|V^{(s)}_{\{L,L'\}}\right|} \sum_{u \in V^{(s)}_{\{L,L'\}}} \mathbbm{1} \left[ L(u) \ne L'(u) \right].
\end{equation}

\subsubsection{Geometric Change}
We can define a geometric change measure $\zeta_g$ based on the average per-pixel difference between a depth rendering of each model. Let $D = \mathcal{R}_D(\mathcal{M},P,K)$ and $D' = \mathcal{R}_D(\mathcal{M}',P,K)$ be depth renderings of the two models from $P$, and $V^{(d)}_\Delta$ be the set of pixels that have a valid depth value for all $D'' \in \Delta$, with $V^{(d)}_D \equiv V^{(d)}_{\{D\}}$. Then we can define
\begin{equation}
\textstyle
\zeta_g = \frac{1}{\left|V^{(d)}_{\{D,D'\}}\right|} \sum_{u \in V^{(d)}_{\{D,D'\}}} \left\|D(u) - D'(u)\right\|_2.
\end{equation}
We report $\zeta_g$ as a value in millimeters. Note that this measure would be particularly high for the depth renderings from pose $P_j$ in Fig.\ \ref{fig:rendering}, since in one of the models, a door has been moved so as to block the view.

\subsubsection{Change Statistics} Please note that change statistics for each scene can be found in the supplementary material.

\subsection{Measuring Re-Localization Performance}
\label{sec:evaluation_metrics}

Given a sequence of ground truth poses as 3D orientations $\{{R}_{1}, ..., {R}_p\}$ (where ${R}_{i} \in \mathbb{R}^{3 \times 3}$) and absolute 3D locations $\{t_1, ..., t_p\}$ (with ${t}_{i} \in \mathbb{R}^3$), as well as corresponding pose estimates $\{\hat{{R}}_1, ..., \hat{{R}}_p\}$ and $\{\hat{t}_1, ..., \hat{t}_p\}$, common evaluation protocols are based on absolute pose errors. More specifically, it is common to compute the absolute translation error as a Euclidean distance in meters, namely $\Delta{}t_i = ||\hat{t}_i - t_i||$, and the absolute orientation error as an angle in degrees, namely $\Delta{}\theta_i = ||\frac{180}{\pi} \cdot 2 \cdot \arccos[q(R_i)^{-1} \cdot q(\hat{R}_i)]||$, in which $q(R)$ denotes the quaternion corresponding to the rotation matrix $R$.
Methods can then be ranked by comparing their values for $\mathcal{E}_a$ or $\bar{\mathcal{E}}_a$, the fraction of images localized within (eq. \ref{eq:recall_threshold}) or outside of (eq. \ref{eq:recall_threshold_outlier}) the given error thresholds \mbox{$(\epsilon_{t}, \epsilon_{\theta})$}: 

\begin{equation}
\textstyle
\label{eq:recall_threshold}
        \mathcal{E}_a(\epsilon_t, \epsilon_{\theta}) = \frac{1}{p}\sum_{i=1}^p \mathbbm{1} \left[ \Delta{}t_i < \epsilon_t \text{ and } \Delta{}\theta_{i} < \epsilon_{\theta} \right]
\end{equation}

\begin{equation}
\textstyle
\label{eq:recall_threshold_outlier}
        \bar{\mathcal{E}}_a(\epsilon_t, \epsilon_{\theta}) = \frac{1}{p}\sum_{i=1}^p \mathbbm{1} \left[ \Delta{}t_i \geq \epsilon_t \text{ or } \Delta{}\theta_{i} \geq \epsilon_{\theta} \right]
\end{equation}

\noindent Commonly chosen thresholds for $\mathcal{E}_a$ in indoor setups are $(0.05m, 5^{\circ})$ or $(0.1m, 10^{\circ})$. However, these values are manually selected, and do not correlate with the visual appearance of a scene: a one-pixel shift could potentially lead to a pose error of only a few millimeters when objects are close, but a few meters if objects are far from the camera.
Instead of using hard thresholds, \cite{Kendall2015} independently reports the medians $\widetilde{\Delta{}t}$ and $\widetilde{\Delta{}\theta}$ of the absolute translation and angular errors. However, these median errors can correspond to completely different frames, and there is in fact no guarantee with this measure that any single frame has both a low translation error and a low angular error, even if both the medians are low. In this paper, we eschew both of these approaches and instead propose a new measure that, rather than being based on the absolute translation and angular errors, is directly based on the difference in appearance between an image from the ground truth pose and an image from the predicted pose.

\begin{figure}[!t]
    \centering
    \includegraphics[width=\linewidth]{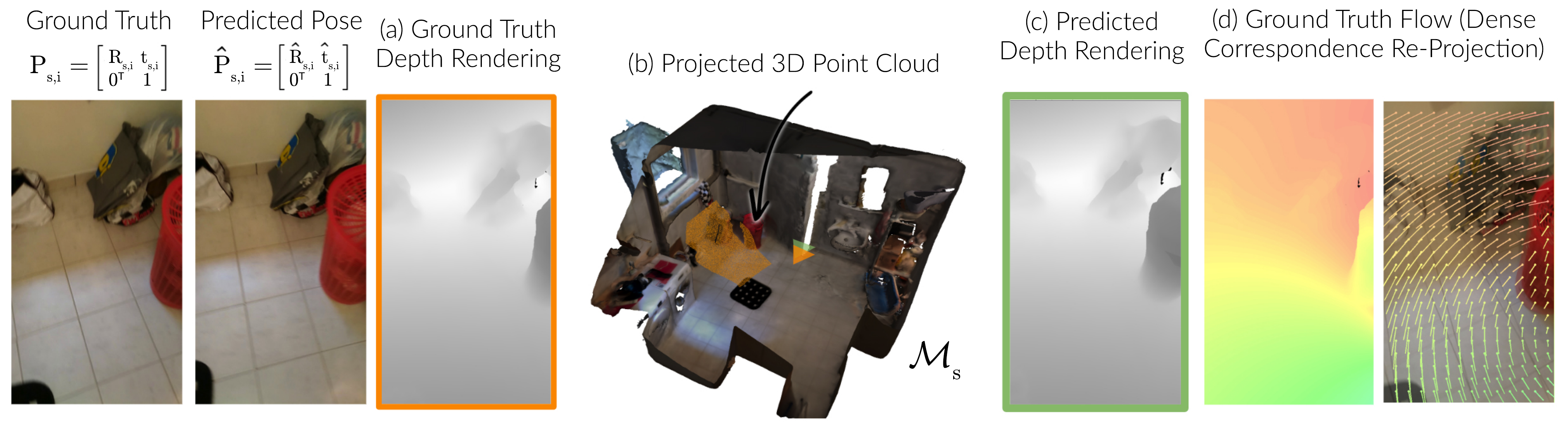} 
    \caption{Given ground truth and predicted camera poses ${P}_{s,i}$ and $\hat{{P}}_{s,i}$, we compute the flow errors $\delta_f^{(i)}(u)$ for DCRE by back-projecting the rendered depth image $D = \mathcal{R}_D(\mathcal{M}_s,P_{s,i},K_s)$ in (a) using $\Pi_{K_s}^{-1}$ to get a 3D point cloud (b) that is then transformed by ${\hat{P}}_{s,i}^{-1}{P}_{s,i}$. The flow errors are the displacements between the projections of the points in this transformed point cloud and the pixels in the original image. See Equation~\ref{eq:flow_xy}.}
    \label{fig:optical_flow} 
\end{figure}

\subsubsection{\nameEvalMetricLong{} (\nameEvalMetricShort)}
\label{sec:reprojection_error}

Our new measure, which we call the \textit{\nameEvalMetricLong}, is defined as a ground truth re-projection error of the 2D flow of dense 3D points rendered from an underlying 3D model (see Fig. \ref{fig:optical_flow}).
The flow is computed according to our ground truth and predicted camera poses. Specifically, the 3D model for the sequence of interest $s$ is first rendered from the ground truth pose ${P}_{s,i}$. This gives us a high-resolution dense depth map $D_i = \mathcal{R}_D(\mathcal{M}_s, {P}_{s,i}, K_s)$ that can be back-projected into a 3D point cloud using the back-projection function $\Pi_{K_s}^{-1}$.
The points in the cloud are then transformed by ${\hat{P}}_{s,i}^{-1}{P}_{s,i}$ before being projected back down onto the image plane using $\Pi_{K_s}$ to get a new depth map. The flow error $\delta_f^{(i)}$ at a pixel $u$ in frame $i$ can then be defined as

\begin{equation}
    \label{eq:flow_xy}
     \delta_f^{(i)}(u) = \Pi_{K_s}(\hat{P}^{-1}_{s,i} P_{s,i} \Pi_{K_s}^{-1} (u, D_i)) - u.
\end{equation}

\noindent
Intuitively, the overall frame error $\mathcal{E}_{\nameEvalMetricShort}^{(i)}$ is then the average magnitude of the 2D correspondence displacement, normalised by the image diagonal, \ie{}

\begin{equation}
    \label{eq:flow_overall}
    \mathcal{E}_{\nameEvalMetricShort}^{(i)} = \frac{1}{\left|V^{(d)}_{D_i}\right|} \sum_{u \in V^{(d)}_{D_i}} \min\left(\frac{||\delta_f^{(i)}(u)||}{\sqrt{w^2 + h^2}}, 1\right).
\end{equation}
This can then be extended to a \nameEvalMetricShort{}-based error $\mathcal{E}_f(\epsilon_f)$ for the whole sequence:

\begin{equation}
\textstyle
\label{eq:recall_flow}
        \mathcal{E}_f(\epsilon_f) = \frac{1}{p}\sum_{i=1}^{p} \mathbbm{1} \left[ \mathcal{E}_{\nameEvalMetricShort}^{(i)} < \epsilon_f \right].
\end{equation}

\begin{figure}[!t]
    \centering
  \includegraphics[width=\linewidth]{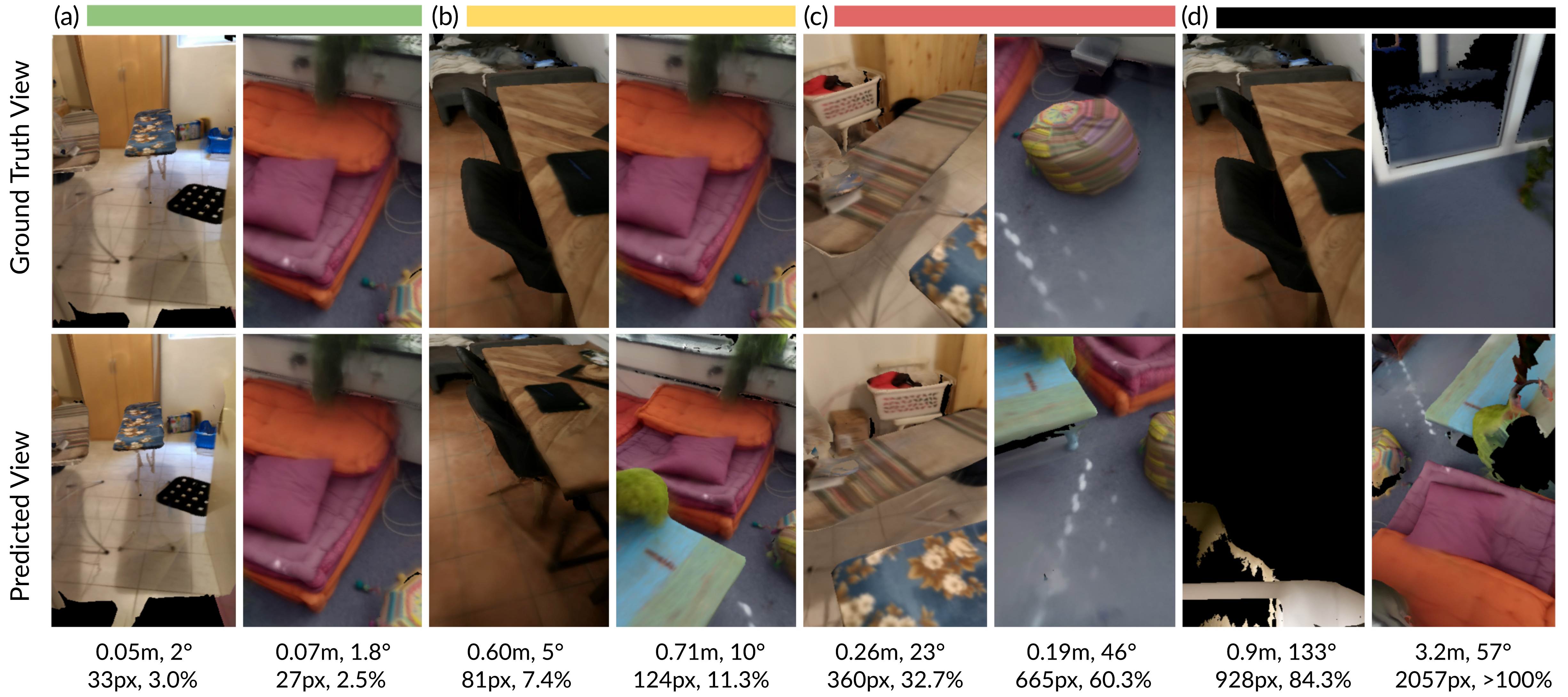} 
  \caption{Some example poses predicted by the methods evaluated in Sec.\ \ref{sec:reprojection_error}, their absolute pose errors in m/$^\circ$, and the \nameEvalMetricShort{} in pixels and the percentage of the image diagonal this represents in each case. See also Figs.\ \ref{fig:cumulative_overview}, \ref{fig:change_correlation} and \ref{fig:hist_multiframe}. 
    }
    \label{fig:dcre_examples} 
\end{figure}

\noindent One major advantage of such a measure is that it gives us an error that correlates with visual perception (see examples in Fig. \ref{fig:dcre_examples}). Another desirable property of \nameEvalMetricShort{} is the fact that it is represented by a single number, which is in contrast to absolute pose errors, which struggle to combine the translation error $\Delta{}t_i$ with the angular error $\Delta{}\theta_i$. Furthermore, a (cumulative) \nameEvalMetricShort{} histogram can provide us with a good way of characterising the performance of a method (see Sec. \ref{sec:results}), since it represents the poses within a wide error range.

\section{Experiments}

To evaluate the impact of appearance changes on indoor camera re-localization, we analyse the performance of state-of-the-art re-localizers on \emph{\nameDataset{}} using both common evaluation metrics and our newly proposed DCRE measure (Sec. \ref{sec:evaluation_metrics}). We also conduct experiments to evaluate how robust different re-localizers are with respect to various types of change, as suggested in Sec. \ref{sec:classify_change}. 

\subsection{Classifying Frame Difficulty}
\label{sec:frame_difficulty}

Scene changes are one factor that can make single-image re-localization challenging, but other factors (\eg{} scene context and texture, or the pose novelty with respect to the training trajectory) can also play a significant role. We thus propose to rank the difficulty of each query image based on the following three properties. More details can be found in the supplementary material.

\PAR{Variance of Laplacian} Many feature-based methods struggle when confronted with motion blur and a lack of texture. To be able to detect such images, we compute the variance of the Laplacian of the image, which we refer to as $\sigma$.

\begin{figure}[!t]
    \centering
 \includegraphics[width=0.93\linewidth]{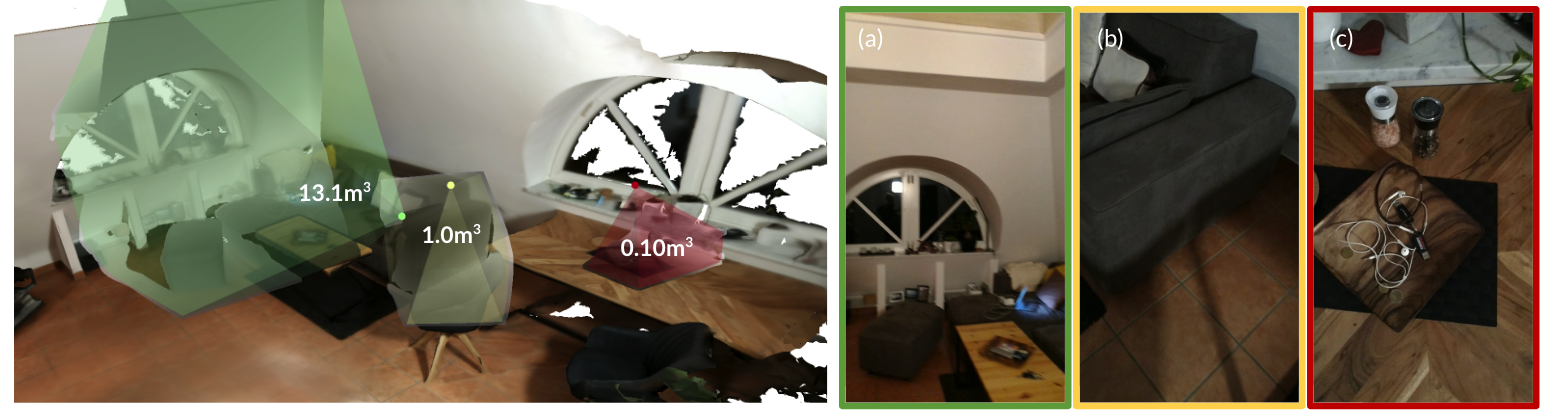} 
   \caption{(a) High $13.1m^3$, (b) medium $1.0m^3$ and (c) low $0.1m^3$ frame coverage.}
    \label{fig:frame_coverage} 
\end{figure}

\PAR{Field of View Context} Besides a lack of texture, a lack of scene context can present another major challenge for camera re-localizers. To estimate the field of view of a particular frame, we first back-project the depth map. The volume of the convex hull of the resulting 3D points, combined with the camera center, gives an estimate of the context observed in a particular view (see Fig. \ref{fig:frame_coverage}).

\PAR{Pose Novelty} Another major challenge for camera re-localizers is the novelty of query poses with respect to the training trajectory. Given a sequence of poses $\{P'_{0}, ..., P'_{p}\}$ from the train set, and a ground truth query pose $P$, we can define the pose novelty $\eta$ as the minimum of some dissimilarity function $\epsilon_\eta$ between all pose combinations, such that $\eta = \min_{\forall P'_{i} \in \{P'_{0}, ..., P'_{p}\}} \epsilon_\eta(P, P'_{i})$.

\subsection{Re-Localization Performance} 
\label{sec:results}

\begin{figure}[!t]
    \centering
    \includegraphics[width=\linewidth]{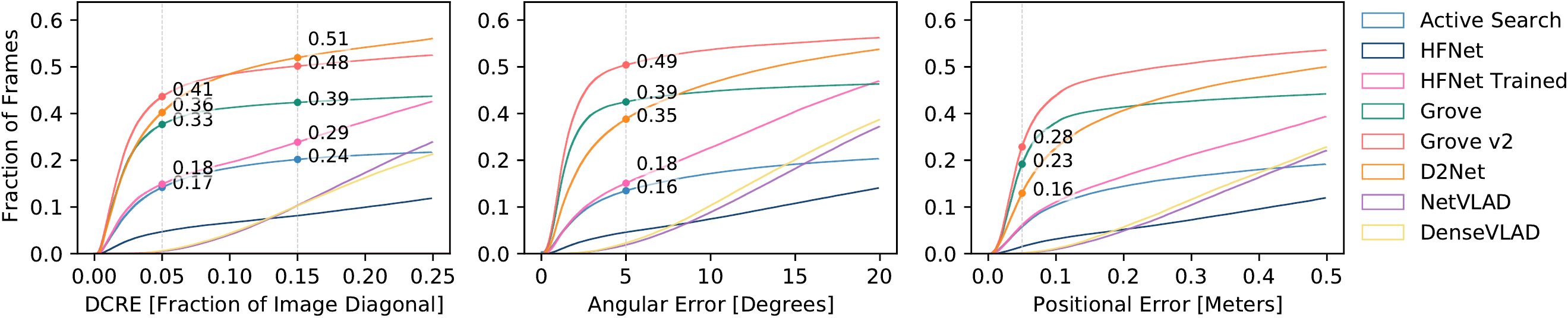} 
    \caption{Cumulative plots of the absolute pose recall and \nameEvalMetricShort{} for all camera re-localization methods.}
    \label{fig:cumulative_overview} 
\end{figure}

In the following, we evaluate a selection of state-of-the-art algorithms that cover the most common types of re-localization approach: hand-crafted structure-based methods \cite{Sattler2011}, learned methods\footnote{Training details for HF-Net can be found in the supplementary material. \label{hfnet_details}} that expect either RGB \cite{Sarlin2019coarse,Dusmanu2019CVPR} or RGB-D \cite{Cavallari2017,Cavallari2019PAMI} input, and image retrieval methods \cite{Torii2015,Arandjelovic2016}. In our first experiment, we evaluate on all \num{165744} query test images, without any filtering. We list the overall performance of each method in Tbl.~\ref{table:relocalization_overview} by reporting $\mathcal{E}_f(0.05)$ and $\mathcal{E}_f(0.15)$, based on our newly introduced \nameEvalMetricShort{} metric. For comparison, we also report the recall based on the absolute pose error $\mathcal{E}_a$, with the often-used thresholds $(\epsilon_t, \epsilon_\theta) = (0.05m, 5^{\circ})$.
Further, we also quantify the number of re-localization outliers\footnote{We define $\bar{\mathcal{E}}_f(\epsilon_f) = \frac{1}{p}\sum_{i=1}^{p} \mathbbm{1} \left[ \mathcal{E}_{\nameEvalMetricShort}^{(i)} \ge \epsilon_f \right].$ \label{outlier_eq}}, by reporting both the percentage of frames with a high $\mathcal{E}_a$ error, with $(\epsilon_t, \epsilon_\theta) = (0.5m, 25^{\circ})$, high \nameEvalMetricShort{} error, with $\mathcal{E}_f(0.5)$, and failed re-localizations (no predicted pose or \texttt{NaN}).
Single numbers are still not really descriptive of the dynamics of each algorithm.
We thus visualize cumulative plots in Fig.\ \ref{fig:cumulative_overview} using \nameEvalMetricShort, as well as $\Delta{}t_i$ and $\Delta{}\theta_i$ for comparison.
These graphs shed some light on the behavior of the methods that we analyze. For example, it is interesting that the best-performing methods according to the threshold-based metrics $\mathcal{E}_a$ and $\mathcal{E}_f$, such as \grovetwo{} and D2-Net, output increasingly inaccurate poses, as evidenced by the steady increase in their \nameEvalMetricShort{} values towards the right of the plot. By contrast, Active Search tends to provide poses for a smaller number of query frames but, crucially, does not output overly incorrect poses, as evidenced by the plateauing of its \nameEvalMetricShort{} plot.
While some of this information can also be gained by analysing the numbers in Tbl.~\ref{table:relocalization_overview}, we find that the cumulative plot provides a deeper, more intuitive characterisation of each method. An ideal method should yield a cumulative \nameEvalMetricShort{} that is as similar to a step function as possible: first rising quickly to correctly re-localize a good fraction of the frames, and then plateauing (signalling failed re-localizations instead of producing highly incorrect poses).

\begin{table}[!t]
\centering
\caption{Comparison of all methods w.r.t.\ their inlier/outlier ratios, median pose errors and \nameEvalMetricShort{ }errors. \textit{Obj.} is the fraction of failure cases where the methods re-localized against a moved object. \textit{N/A} denotes invalid/missing predictions.}
\label{table:relocalization_overview}
\resizebox{\textwidth}{!}{
{\renewcommand{\arraystretch}{1.3}
\begin{tabular}{lcccccccc}
\toprule
& \multicolumn{4}{c}{\textbf{Inlier}} & \multicolumn{3}{c}{\textbf{Outlier}} \\
& \multicolumn{4}{c}{\raisebox{0.5ex}{\smash{$\overbrace{\makebox[6.2cm]{}}_{}$}}} & \multicolumn{3}{c}{\raisebox{0.5ex}{\smash{$\overbrace{\makebox[4.5cm]{}}_{}$}}} \\
\textbf{Method} & 
$\mathbf{\mathcal{E}_a(0.05m, 5^{\circ})}$ & 
$(\mathbf{\widetilde{\Delta{}t}}, \mathbf{\widetilde{\Delta{}\theta}})$ &
$\;\mathbf{\mathcal{E}_f(0.05)}\;$ &
$\;\mathbf{\mathcal{E}_f(0.15)}\;$ &
$\;$\textbf{N/A}$\;$ & 
$\mathbf{\bar{\mathcal{E}}_a(0.5m, 25^{\circ})}\quad$ & $\mathbf{\bar{\mathcal{E}}_f(0.5)}$\textsuperscript{\ref{outlier_eq}} & \textbf{ Obj.}\\
\midrule
Active Search \cite{Sattler2017} & 0.0696 & $(0.16, 4.68)$ & 0.171 & \multicolumn{1}{p{1cm}|}{$\;$ 0.243 }  & 0.684 & 0.0891 & 0.028& \multicolumn{1}{|c}{$\;$ 0.149}\\
Grove \cite{Cavallari2017} & 0.2300 & $(0.06, 1.74)$  & 0.334 & \multicolumn{1}{p{1cm}|}{$\;$ 0.391 }  & 0.452 & 0.144 & 0.106 & \multicolumn{1}{|c}{$\;$ 0.065}\\
\grovetwo~\cite{Cavallari2019PAMI} & 0.2742 & $(0.11, 2.60)$  & 0.406 & \multicolumn{1}{p{1cm}|}{$\;$ 0.485 }  & 0.162 & 0.332 & 0.262 & \multicolumn{1}{|c}{$\;$ 0.051}\\
HFNet \cite{Sarlin2019coarse} & 0.0182 & $(1.56, 72.33)$  & 0.057 & \multicolumn{1}{p{1cm}|}{$\;$ 0.098 }  & 0 & 0.900 & 0.714 & \multicolumn{1}{|c}{$\;$ 0.005}\\
HF-Net Trained\textsuperscript{\ref{hfnet_details}} \cite{Sarlin2019coarse} & 0.0725 & $(0.84, 24.17)$  & 0.180 & \multicolumn{1}{p{1cm}|}{$\;$ 0.288 }  & 0 & 0.685 & 0.427 & \multicolumn{1}{|c}{$\;$ 0.065} \\
D2Net \cite{Dusmanu2019CVPR} & 0.1553 & $(0.55, 14.90)$ & 0.365 & \multicolumn{1}{p{1cm}|}{$\;$ 0.506 }  & 0.014 & 0.513 & 0.194 & \multicolumn{1}{|c}{$\;$ 0.033} \\
NetVLAD \cite{Arandjelovic2016} & 0.0002 & $(0.93, 31.44)$ & 0.006 & \multicolumn{1}{p{1cm}|}{$\;$ 0.125 } & 0 & 0.798 & 0.452 & \multicolumn{1}{|c}{$\;$ 0.016}\\
DenseVLAD \cite{Torii2015} & 0.0003 & $(0.98, 32.26)$ & 0.008 & \multicolumn{1}{p{1cm}|}{$\;$ 0.124 } & 0.006 & 0.772 & 0.520 & \multicolumn{1}{|c}{$\;$ 0.014} \\
\bottomrule
\end{tabular}
}}
\end{table}

\begin{figure}[!t]
    \centering
    \includegraphics[width=\linewidth]{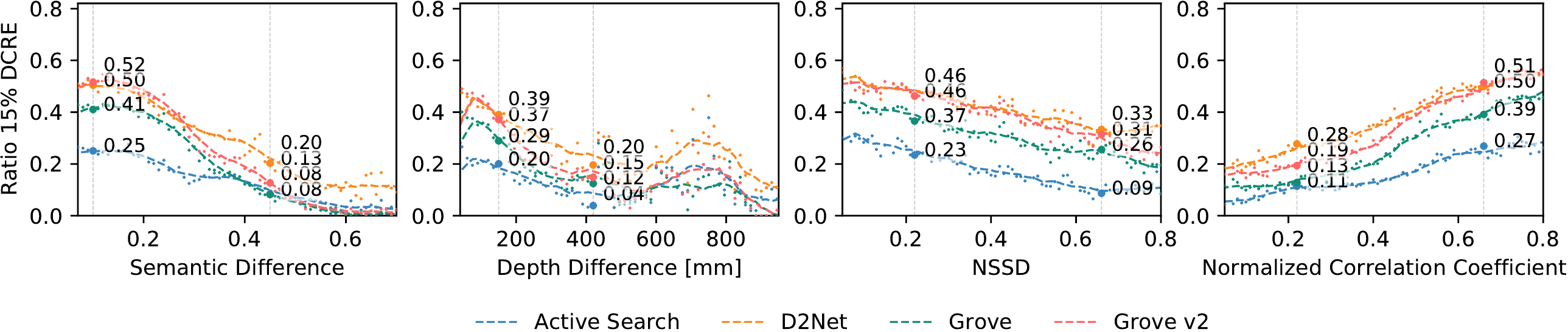}
    \caption{The charts show the performance of the best methods with respect to semantic $\zeta_s$, geometric $\zeta_g$ and visual change ($\zeta_v$ and $\rho_v$). Each dot represents the performance $\mathcal{E}_f(0.15)$ of a particular method on frames with increasing change measured by $\zeta_s$, $\zeta_g$, $\zeta_v$ and $\rho_v$ respectively. Note that the dashed lines denote running averages.}
    \label{fig:change_correlation} 
\end{figure}

\PAR{Scene Changes} To see how scene changes affect a method's performance, we plot the overall error/performance of the best methods with images of increasing visual ($\zeta_v$ and $\rho_v$), geometric  $\zeta_g$ and semantic change $\zeta_s$.
A clear correlation between scene changes and overall performance is observable in Fig.\ \ref{fig:change_correlation}.

\begin{figure}[!t]
    \centering
    \includegraphics[width=0.95\linewidth]{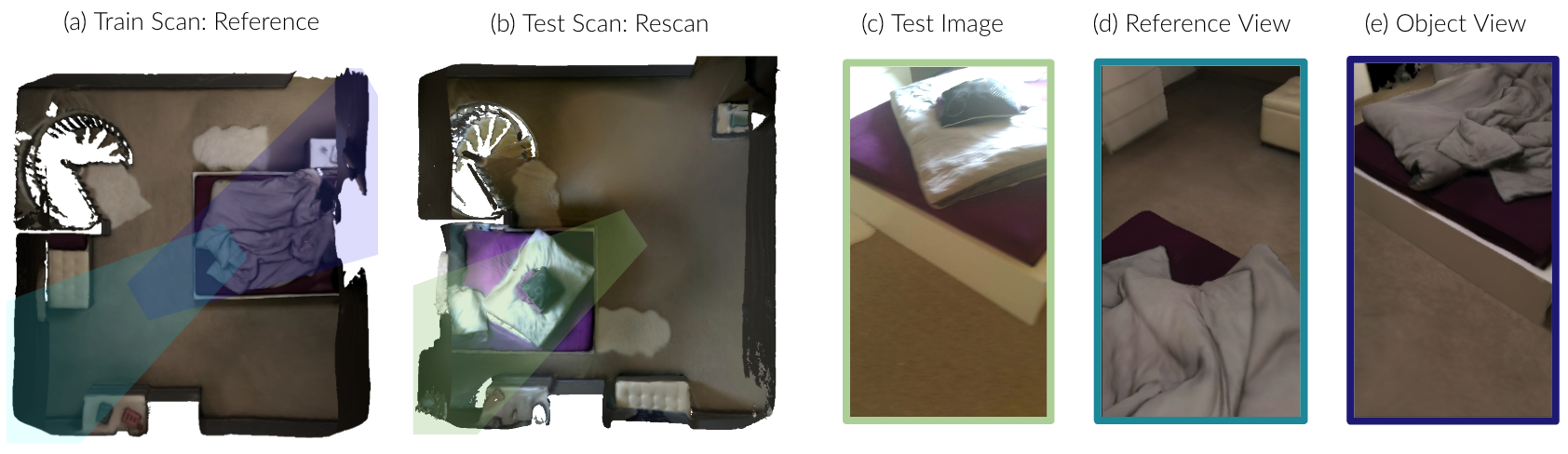} 
    \caption{Given the reference scan (a) for training, localizing the image (c) from a rescan (b) is practically impossible. A camera re-localization method might localize an object (e) instead of the global scene (d).}
    \label{fig:object_relocalization}
\end{figure}

\PAR{Object Re-Localization vs. Camera Re-Localization} Rigidly moving objects cause new types of absolute camera pose estimation ambiguities. Poses become ambiguous when a changed object occupies most of the view. An example is given in Fig. \ref{fig:object_relocalization}, where localizing the test image (c) from the rescan (b) is practically impossible given only the reference scan (a). The correct reference view of the GT pose would produce the reference view pictured in (d). Instead, when an object instance dominates the view, the camera might incorrectly localize with respect to the visible object. We report the fraction of these cases (out of all failure cases) in the last column of Tbl.~\ref{table:relocalization_overview}.

\begin{figure}[!t]
    \centering
    \includegraphics[width=\linewidth]{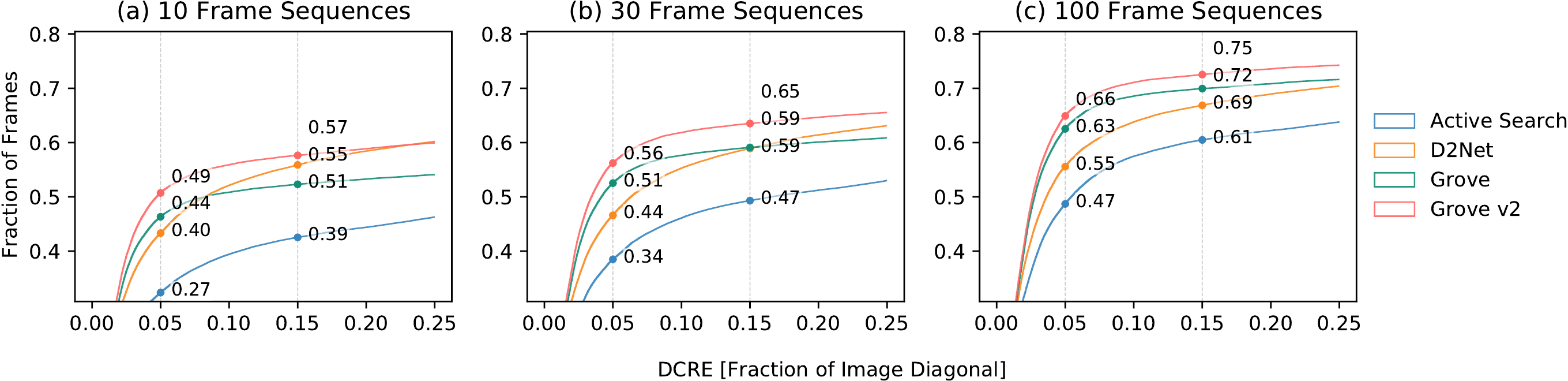} 
    \caption{Cumulative plots of the \nameEvalMetricShort{} for the best-performing camera re-localization methods using (a) short, (b) medium, and (c) long sequences of frames.}
    \label{fig:hist_multiframe} 
\end{figure}

\PAR{Sequences} We experimented with sequence lengths $s_{\Delta}$ of (a) $10$, (b) $30$, and (c) $100$ consecutive frames. The corresponding DCRE plots can be found in Fig. \ref{fig:hist_multiframe}. We chose the values $10$ and $30$ to model interactive applications, where using a small number of consecutive frames can help tackle motion blur and object instance ambiguities; whereas longer sequences of up to $100$ frames can be used in less time-sensitive applications, where re-localization accuracy is more important than interactivity. As the figure shows, when leveraging frame sequences, there is a significant improvement in the DCRE numbers.

\section{Conclusion}
In this paper, we have both curated a suitable dataset for long-term indoor camera re-localization, and defined a set of metrics for quantifying changes in indoor scenes. For the first time, this enables an evaluation of the impact of changes in indoor scenes on re-localization performance, thus closing a significant gap in the literature. We have also introduced \nameEvalMetricShort, a new metric to measure re-localization performance, and shown that many methods experience a loss of performance when exposed to scene regions that have undergone changes of visual, geometric and semantic nature, \eg{} as caused by rigid/non-rigid object movements. We have further analysed the behaviour of camera re-localizers on frames that capture rigidly moving objects. Large semantic changes, \eg{} caused by large objects in a scene changing their position, are a particular problem. 
In such situations, the methods potentially re-localize with respect to the object dominating the camera's field of view, rather than with respect to the scene. Results for state-of-the-art re-localizers on our new benchmark show that none of them is fully capable of handling everyday changes observed in indoor scenes: indeed, there is significant room for improvement. Using short image sequences, rather than individual images, for re-localization naturally improves performance, but is not sufficient to solve our benchmark. We believe that long-term camera re-localization in indoor scenes requires the learning of higher-level concepts of a scene -- such as its semantics, and/or object-level understanding of poses, dynamics and appearance variations -- so as to subsequently be able to reason about scene changes. In this way, we would expect the camera pose estimation task to gradually become more tightly coupled to general scene understanding going forwards.

\section*{Acknowledgements}
This work was supported by the Centre Digitisation Bavaria (ZD.B), the Swedish Foundation for Strategic Research (Semantic Mapping and Visual Navigation for Smart Robots), the Chalmers AI Research Centre (CHAIR) (VisLocLearn), Five AI Ltd.\ and Google Inc.

\clearpage
\section{Supplementary Material}

\noindent This supplementary material provides the following information: 
Sec.~\ref{sec:benchmark} shows the individual scenes of our benchmark dataset. 
Sec.~\ref{sec:change_statistics} provides statistics about the changes that occur in our benchmark dataset (\cf Sec.~4.1 in the main paper). 
Sec.~\ref{sec:difficulties} discusses the means by which we classify the difficulties of the test frames in our dataset (\cf Sec.~5.1 in the main paper). Sec.~\ref{DCRE:metric} provides further details about our DCRE metric (\cf Sec.~4.2 in the main paper). Sec.~\ref{sec:implementation} discusses implementation details for HF-Net and the image retrieval methods we tested (\cf Sec.~5.2 in the main paper), as well as the sequence-based approaches evaluated in the main paper.
In addition to this document, we also provide a supplementary video summarizing our paper.

\subsection{Benchmark Visualization}
\label{sec:benchmark}
Figs.\ \ref{fig:benchmark_overview_s1} -- \ref{fig:benchmark_overview_s10} show the 3D reconstructions of each of the individual scenes in the \emph{\nameDataset} dataset. The scenes selected for \emph{\nameDataset} are very diverse, and exhibit a wide variety of changes, including, but not limited to, complex illumination changes, and appearance variations mostly caused by human interactions, such as rigid object movements (e.g.\ the movement of major objects such as the bed and sofa in scenes 3 and 4, respectively) and non-rigid object deformations (e.g.\ the rearrangement of the blankets in scene 3). Our dataset provides 10 train, 10 validation and 54 test sequences, with \num{52562} images in the train set, \num{34415} images in the validation set and \num{165744} in the test set.

\subsection{Change Statistics}
\label{sec:change_statistics}
\begin{figure}[!h]
    \centering
    \includegraphics[width=\linewidth]{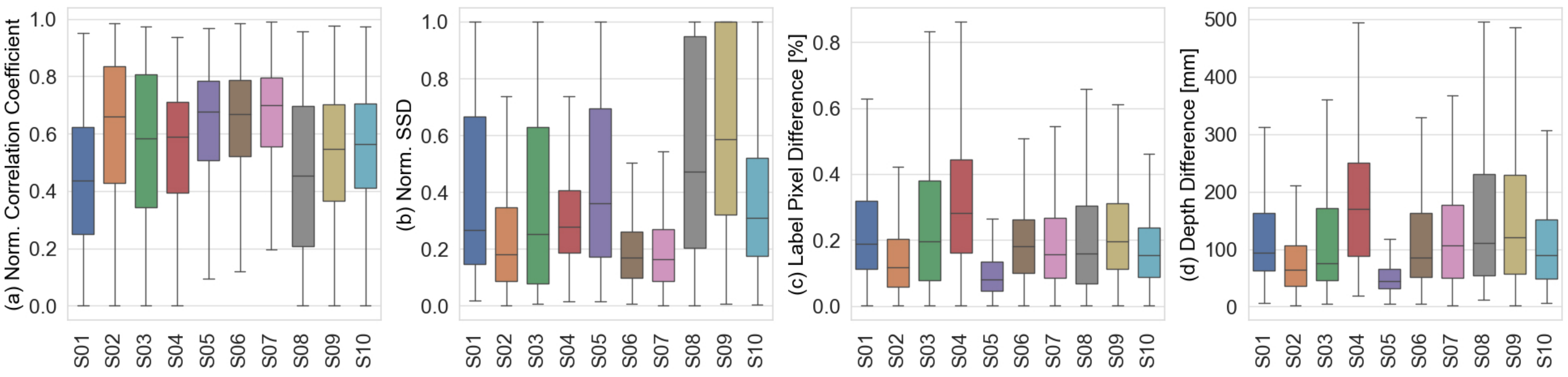} 
    \caption{Visual (a,b), semantic (c) and geometric (d) change statistics for each of the 10 different scenes in our \emph{\nameDataset} dataset. These are computed by averaging the corresponding change measures over all frames from all test sequences for each scene.}
    \label{fig:change_stats}
\end{figure}

Per-scene change statistics corresponding to the change measures described in Sec.\ 4.1 of the main paper can be found in Fig.~\ref{fig:change_stats}. It can be seen that on the one hand, scene 4 has the highest semantic (c) and geometric (d) change values (many objects, including a sofa, move in the rescans), whilst on the other hand, scenes 8 and 9 have a low normalised correlation coefficient~(a) and a high normalised SSD (b), highlighting the visual differences that they contain (see also the original scans in Figs.\ \ref{fig:benchmark_overview_s4}, \ref{fig:benchmark_overview_s8} and \ref{fig:benchmark_overview_s9}).

\subsection{Classifying Frame Difficulty}
\label{sec:difficulties}
\begin{figure}[!t]
    \centering
    \includegraphics[width=0.97\linewidth]{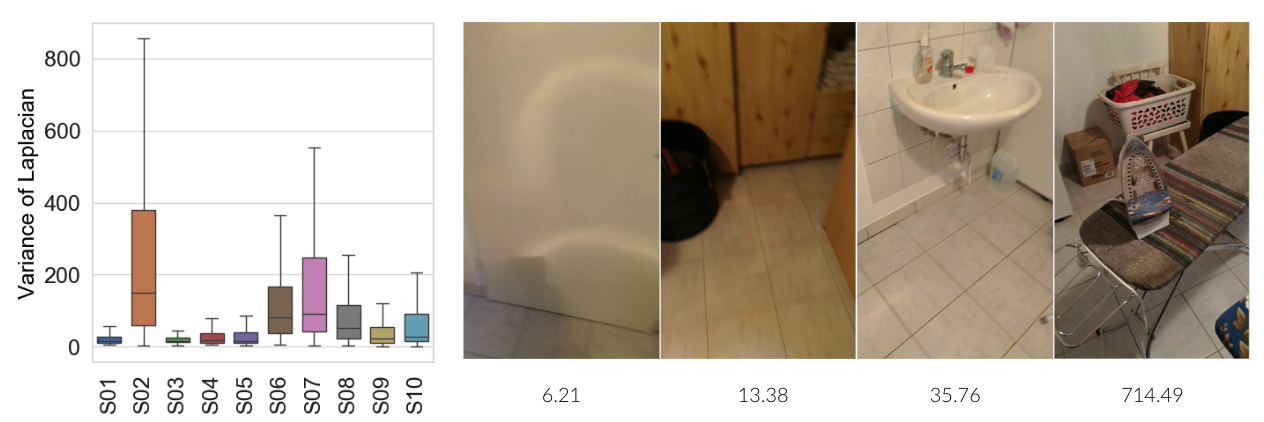} 
    \caption{Left: the average Variance of Laplacian (VoL) value for each test frame, for each scene in our \emph{\nameDataset} dataset. Right: some example test frames from our dataset, and their VoL values. A low VoL value generally indicates that an image exhibits motion blur or a lack of texture (left two images). A high VoL value generally indicates the opposite (right two images).}
    \label{fig:stats_VOL}
\end{figure}

\PAR{Variance of Laplacian (VoL)}
As mentioned in the main paper, the Variance of Laplacian (VoL) measure captures both motion blur and a lack of texture in an image. Fig.~\ref{fig:stats_VOL} shows the average of this measure over all frames from all test sequences for each scene. Blurred images, or images with a lack of texture, such as the left two images in the figure, have a low VoL value and often lack features needed for localization, which can sometimes make them difficult even for humans to localize. By contrast, images with a higher VoL value, such as the right two images in the figure, often contain more descriptive features and are therefore expected to be easier for feature-based re-localization algorithms such as Active Search \cite{Sattler2011} to handle.

\begin{figure}[!t]
    \centering
    \includegraphics[width=\linewidth]{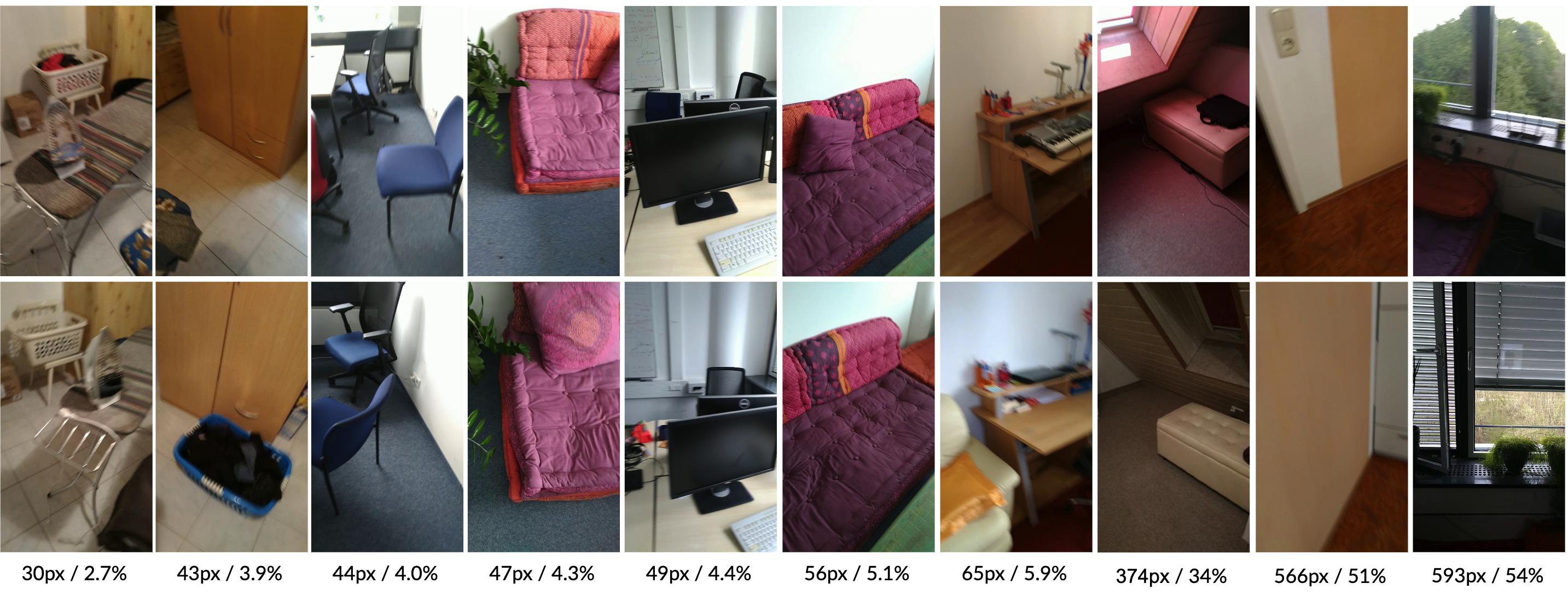} 
    \caption{Visualizing our pose novelty metric. Top row: test images; bottom row: nearest neighbour training images, as computed by using our novel DCRE measure as a pose similarity metric. The DCRE (in pixels), which is used to capture the pose novelty between each pair of images, is printed below them, as is the percentage of the image diagonal it represents in each case.}
    \label{fig:pose_nn}
\end{figure}

\PAR{Pose Novelty} Fig.~\ref{fig:pose_nn} shows a selection of test images, together with their nearest neighbours in the corresponding training sequences, as computed by using our novel DCRE measure as a pose similarity metric. Image pairs with higher DCREs broadly correspond to test images that were captured from more novel poses.

\begin{figure}[!t]
    \centering
    \includegraphics[width=\linewidth]{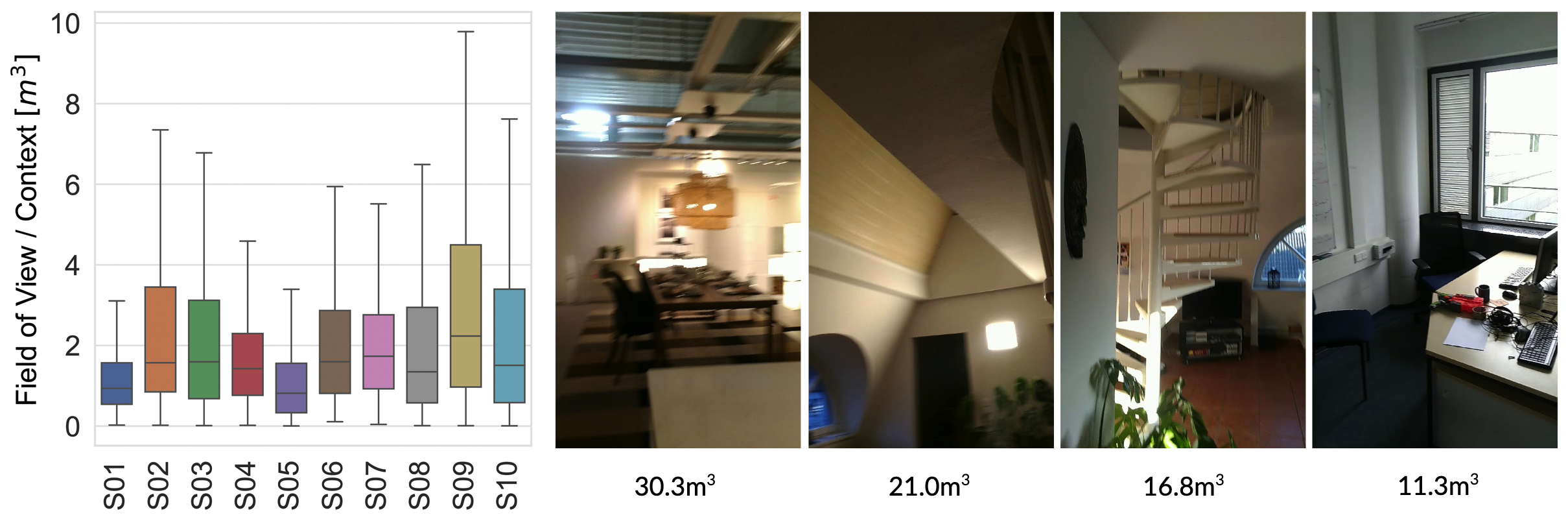} 
    \caption{Visualizing our field of view/context metric. Left: the average field of view/context value for each test frame, for each scene in our \emph{\nameDataset} dataset. Right: some example test frames from our dataset that have particularly high field of view/context values ($> 10m^3$).}
    \label{fig:coverage_stats} 
\end{figure}

\PAR{Field of View/Context}
The left side of Fig.~\ref{fig:coverage_stats} shows the average field of view/context for a test frame in each scene of our \emph{\nameDataset} dataset, as per the description of this metric in Sec.\ 5.1 of the main paper. On the right side of Fig.~\ref{fig:coverage_stats}, some example test images with a context of $> 10m^3$ are shown. In our experiments in Table~\ref{table:performance_filters}, it can be seen that methods struggle with low-context frames (compared to medium-context ones). Interestingly, our high-context frames proved more challenging than our medium-context ones on average, potentially due to a combination of factors such as motion blur, lack of texture and large scene element changes in some of our high-context frames.

\subsection{Dense Correspondence Re-Projection Error}
\label{DCRE:metric}

DCRE is a dense re-projection error of ground-truth correspondences applied in a novel context, namely the evaluation of camera re-localization. Compared to traditional applications (e.g.\ camera calibration or bundle adjustment), synthetic depth images are used, which gives us ground-truth correspondences. The DCRE will thus be 0 for the ground-truth pose. By contrast, the reprojection errors for SfM point clouds will generally be non-zero due to noise in the image measurements. The point clouds generated by rendering a 3D mesh further enable us to compute the metric densely over the whole image rather than being restricted to well-textured regions in the images where features are extracted. Despite its advantages, we are not aware of any re-localization benchmark that uses dense re-projection errors for evaluation.

\PAR{}Please note that we decided to normalize the DCRE error (see eq.\ 9). While normalization is not strictly necessary for this dataset, as all images have the same resolution, it helps future research to compare errors across datasets. 5px is not much in a 5K image, but might be significant at lower image resolutions. 

\subsection{Implementation Details}
\label{sec:implementation}

\begin{figure}[h]
    \centering
    \includegraphics[width=\linewidth]{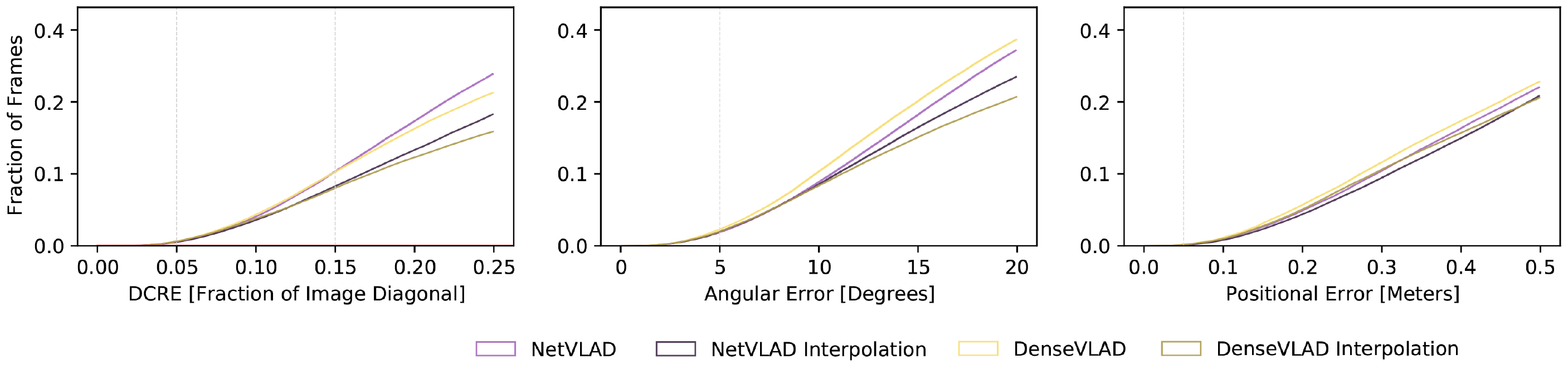} 
    \caption{Cumulative absolute pose recall and DCRE for image retrieval methods.}
    \label{fig:plots_image_retrieval} 
\end{figure}

\begin{figure}[!p]
    \centering
    \includegraphics[width=\linewidth]{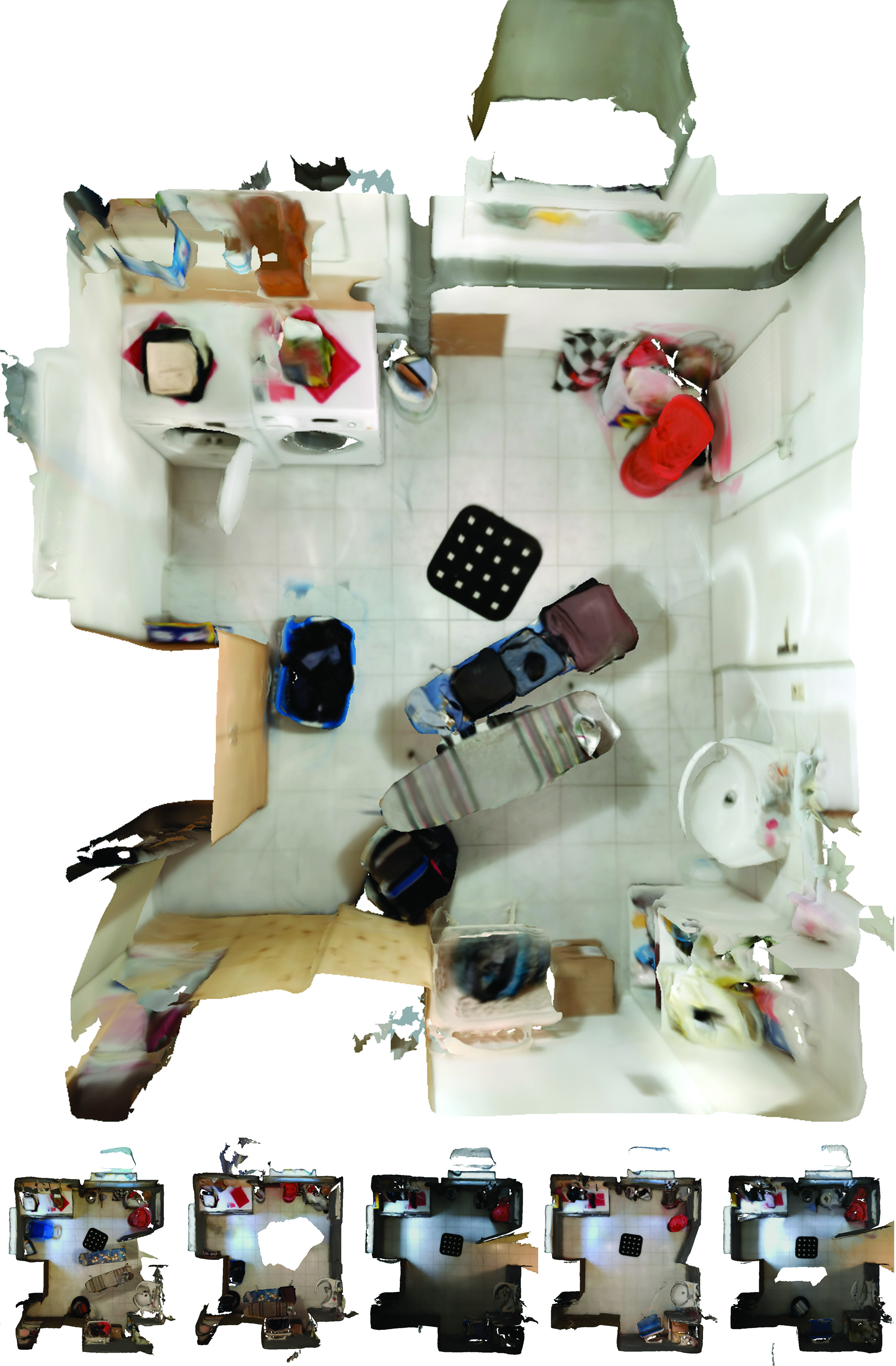} 
    \caption{3D reconstructions of scene 1 of our benchmark dataset.}
    \label{fig:benchmark_overview_s1} 
\end{figure}

\begin{figure}[!p]
    \centering
    \includegraphics[width=\linewidth]{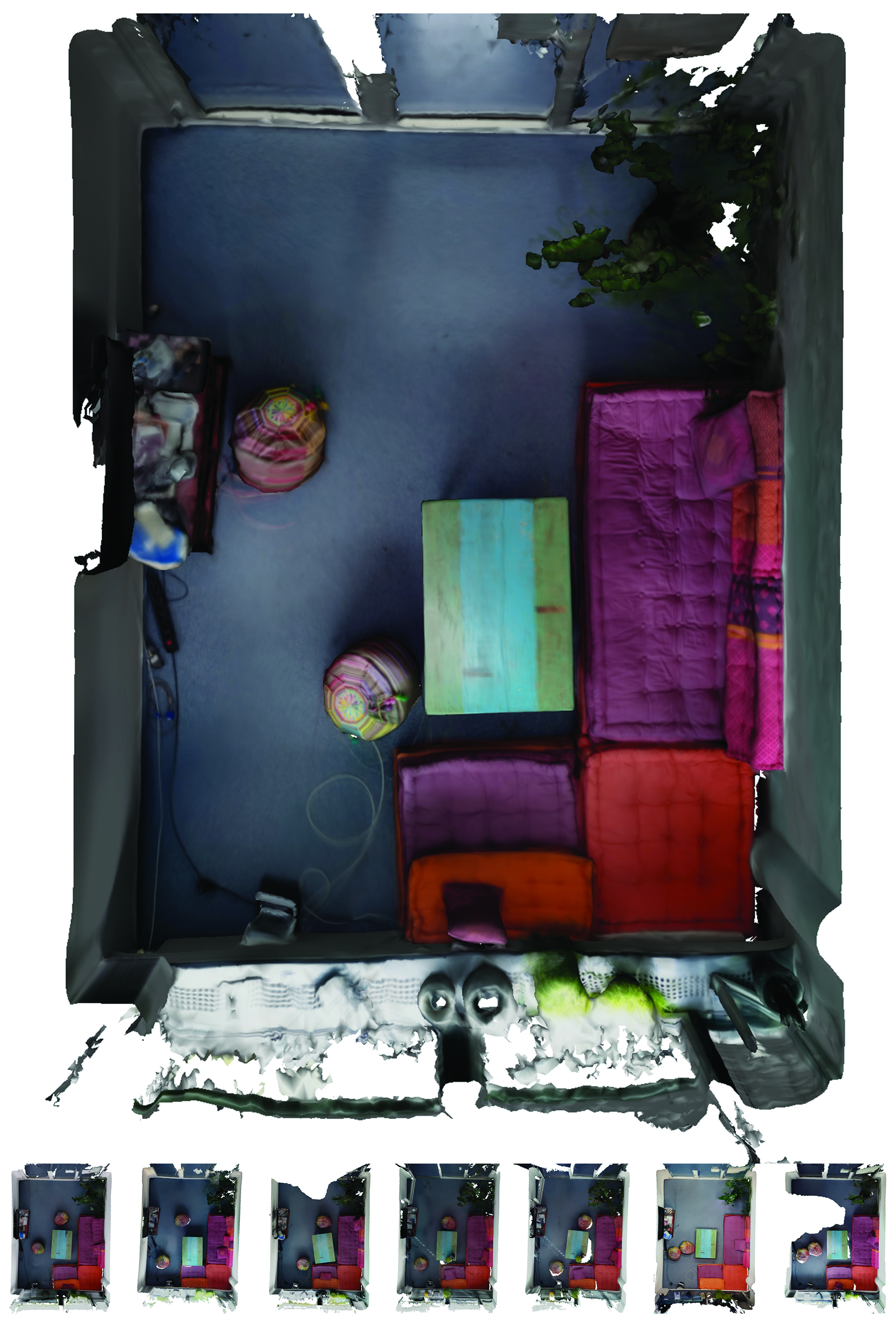} 
    \caption{3D reconstructions of scene 2 of our benchmark dataset.}
    \label{fig:benchmark_overview_s2} 
\end{figure}

\begin{figure}[!p]
    \centering
    \includegraphics[width=\linewidth]{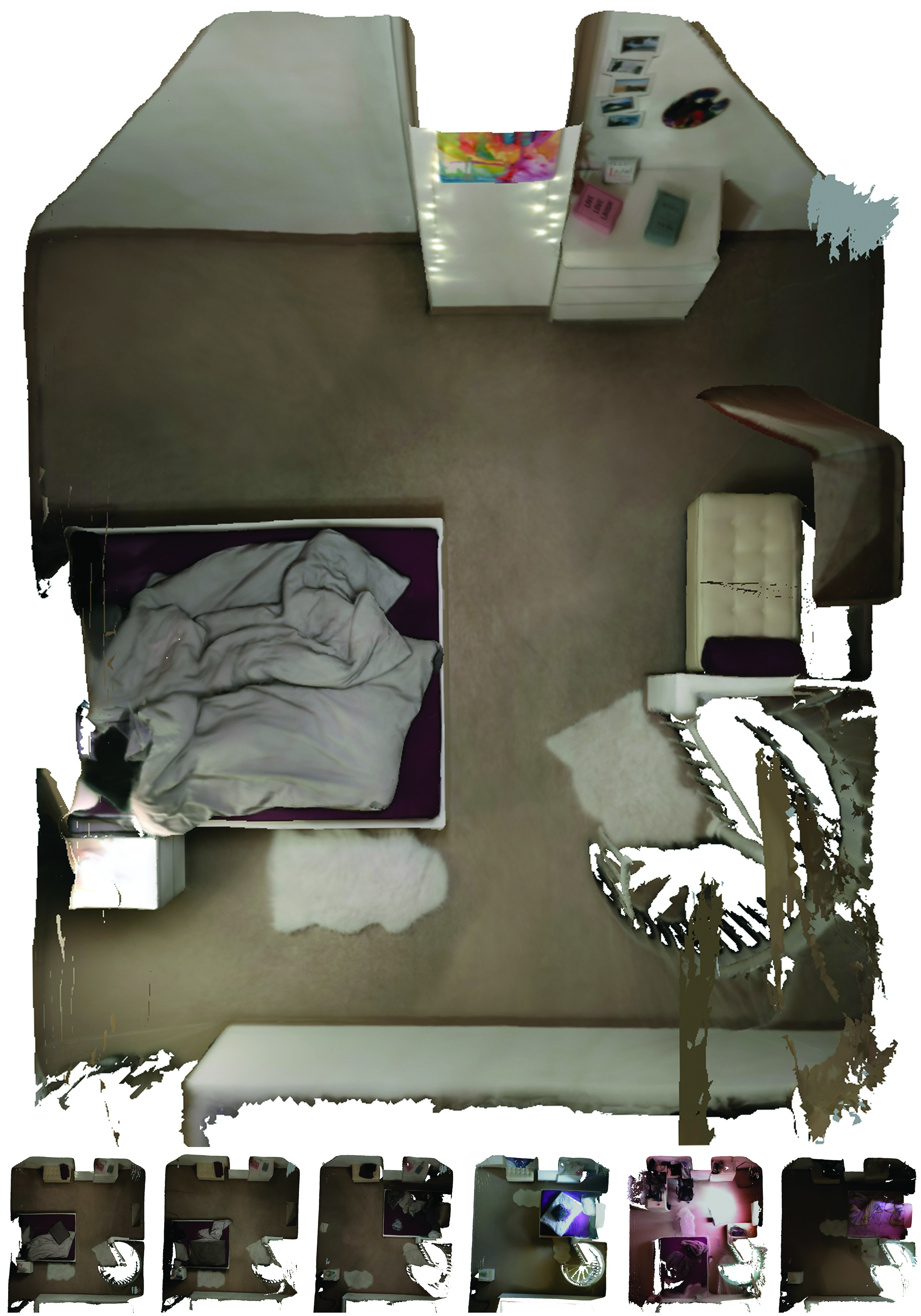} 
    \caption{3D reconstructions of scene 3 of our benchmark dataset.}
    \label{fig:benchmark_overview_s3} 
\end{figure}

\begin{figure}[!p]
    \centering
    \includegraphics[width=\linewidth]{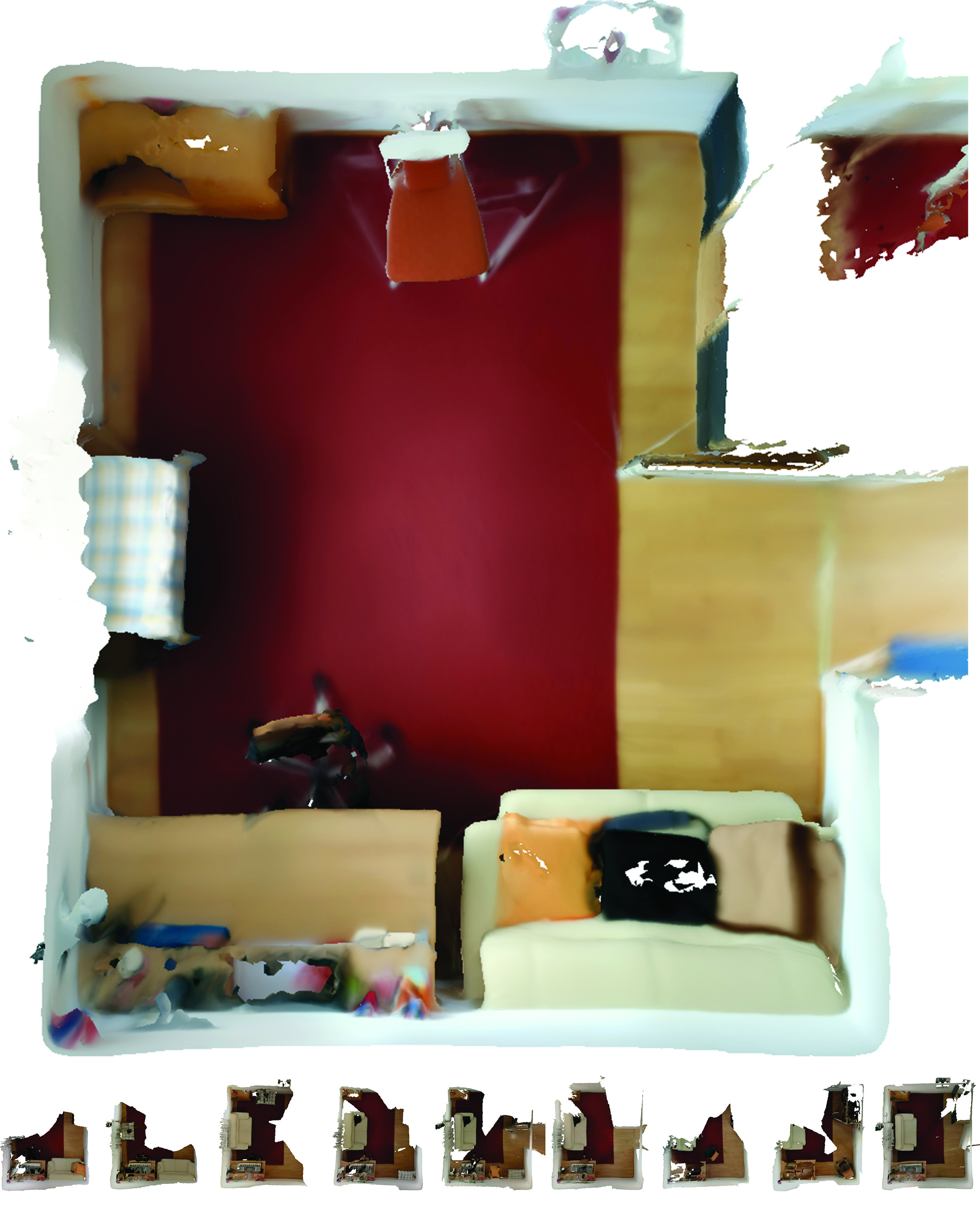} 
    \caption{3D reconstructions of scene 4 of our benchmark dataset.}
    \label{fig:benchmark_overview_s4} 
\end{figure}

\begin{figure}[!p]
    \centering
    \includegraphics[width=\linewidth]{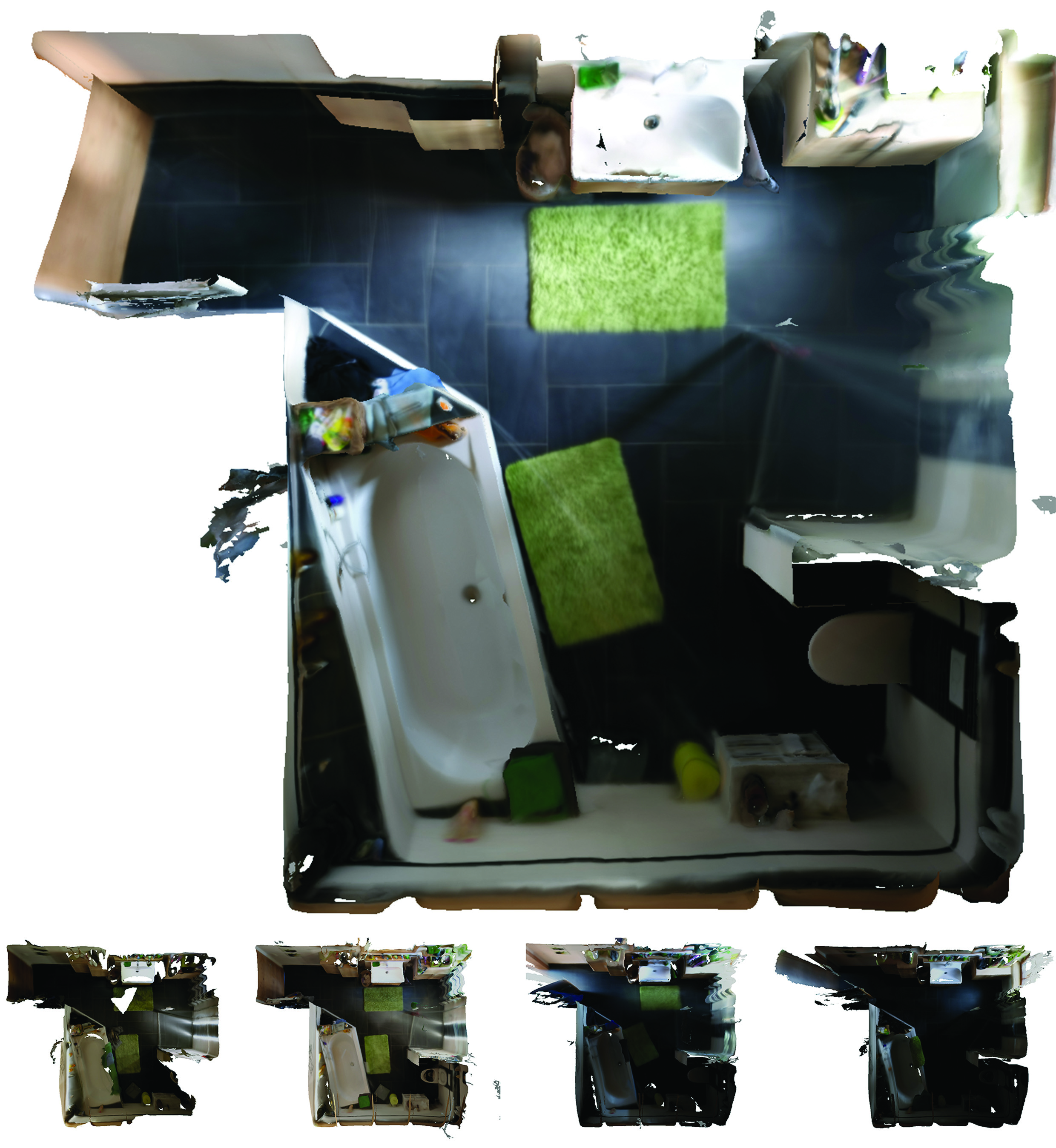} 
    \caption{3D reconstructions of scene 5 of our benchmark dataset.}
    \label{fig:benchmark_overview_s5} 
\end{figure}

\begin{figure}[!p]
    \centering
    \includegraphics[width=\linewidth]{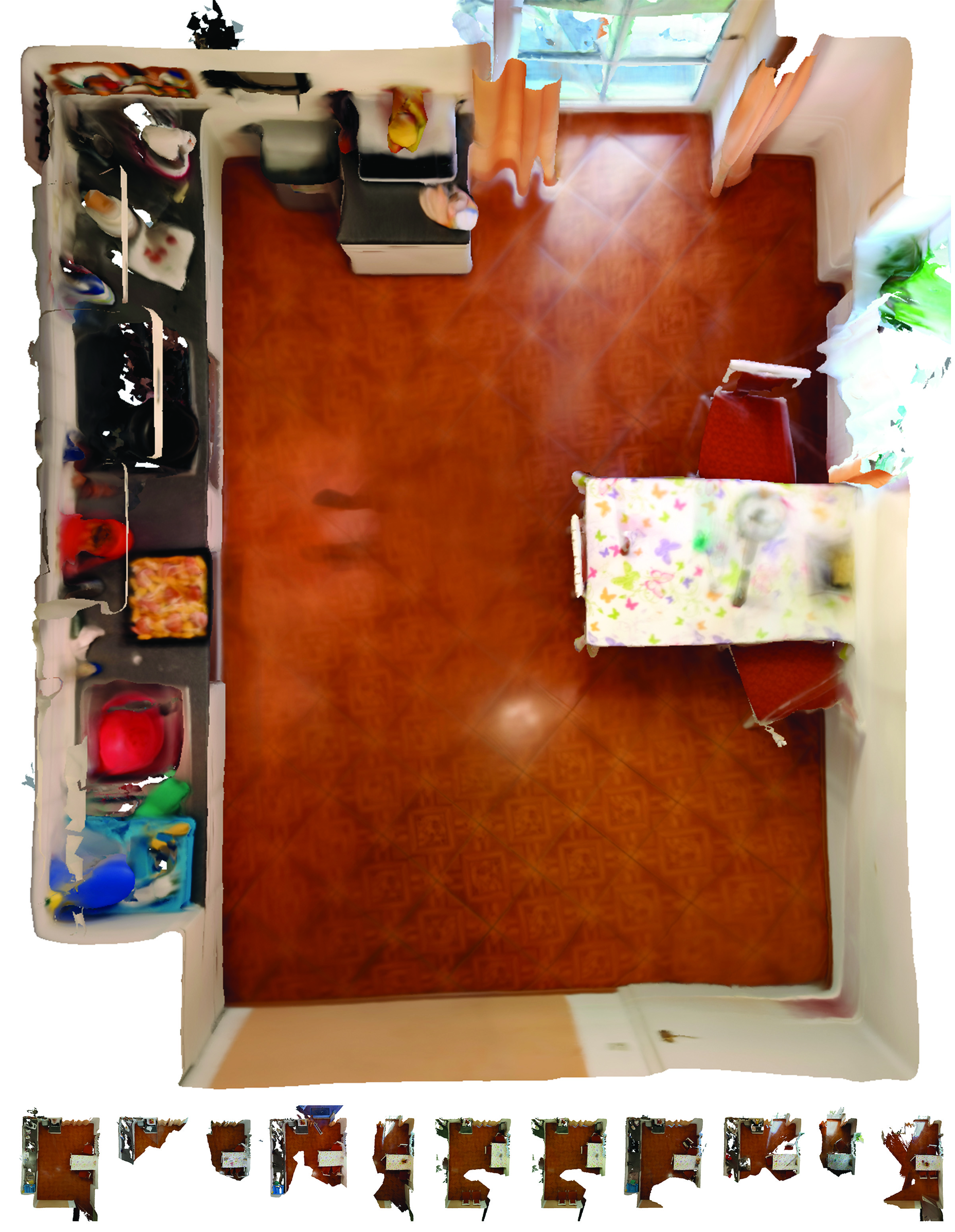} 
    \caption{3D reconstructions of scene 6 of our benchmark dataset.}
    \label{fig:benchmark_overview_s6} 
\end{figure}

\begin{figure}[!p]
    \centering
    \includegraphics[width=\linewidth]{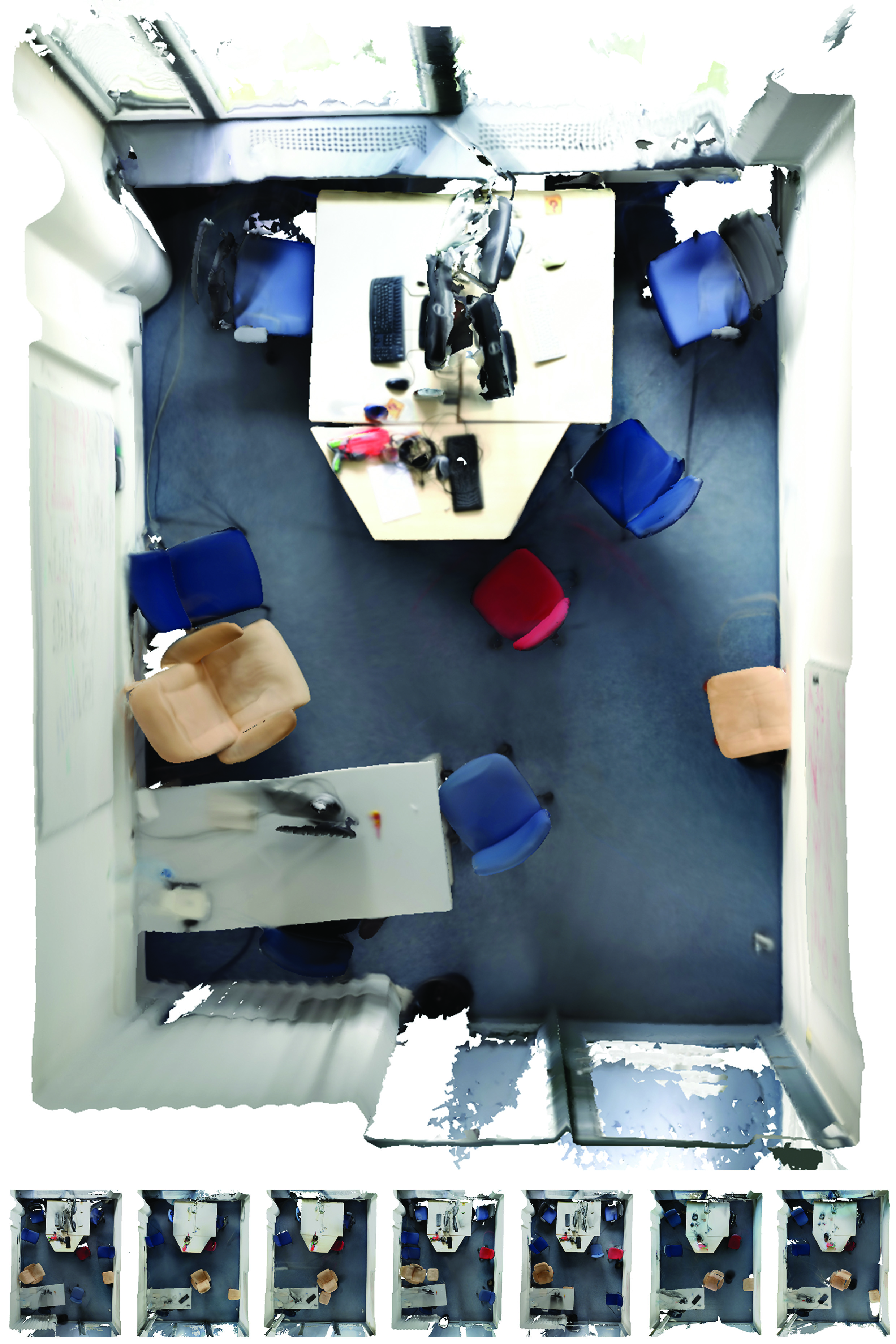} 
    \caption{3D reconstructions of scene 7 of our benchmark dataset.}
    \label{fig:benchmark_overview_s7} 
\end{figure}

\begin{figure}[!p]
    \centering
    \includegraphics[width=\linewidth]{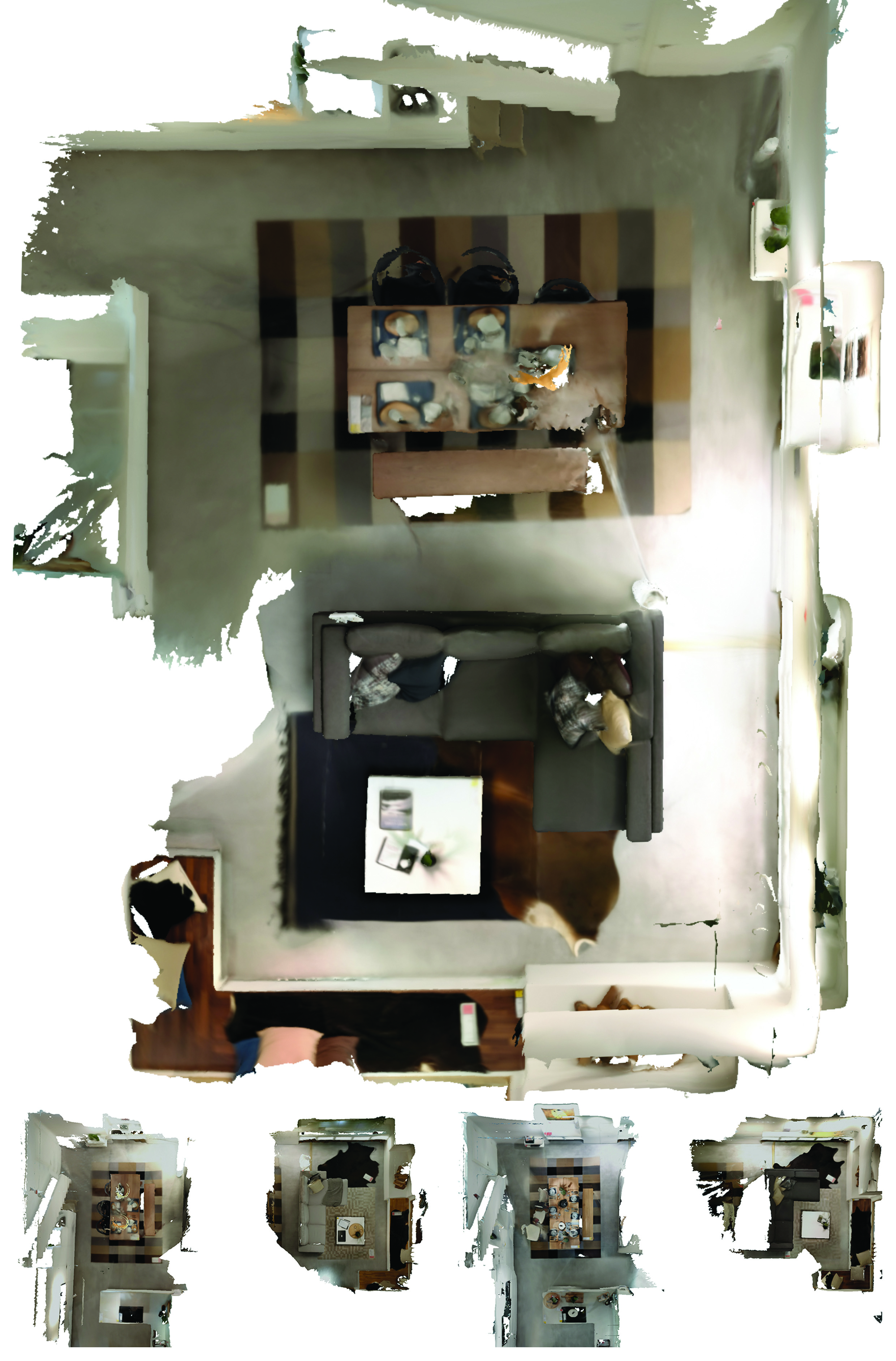} 
    \caption{3D reconstructions of scene 8 of our benchmark dataset.}
    \label{fig:benchmark_overview_s8} 
\end{figure}

\begin{figure}[!p]
    \centering
    \includegraphics[width=\linewidth]{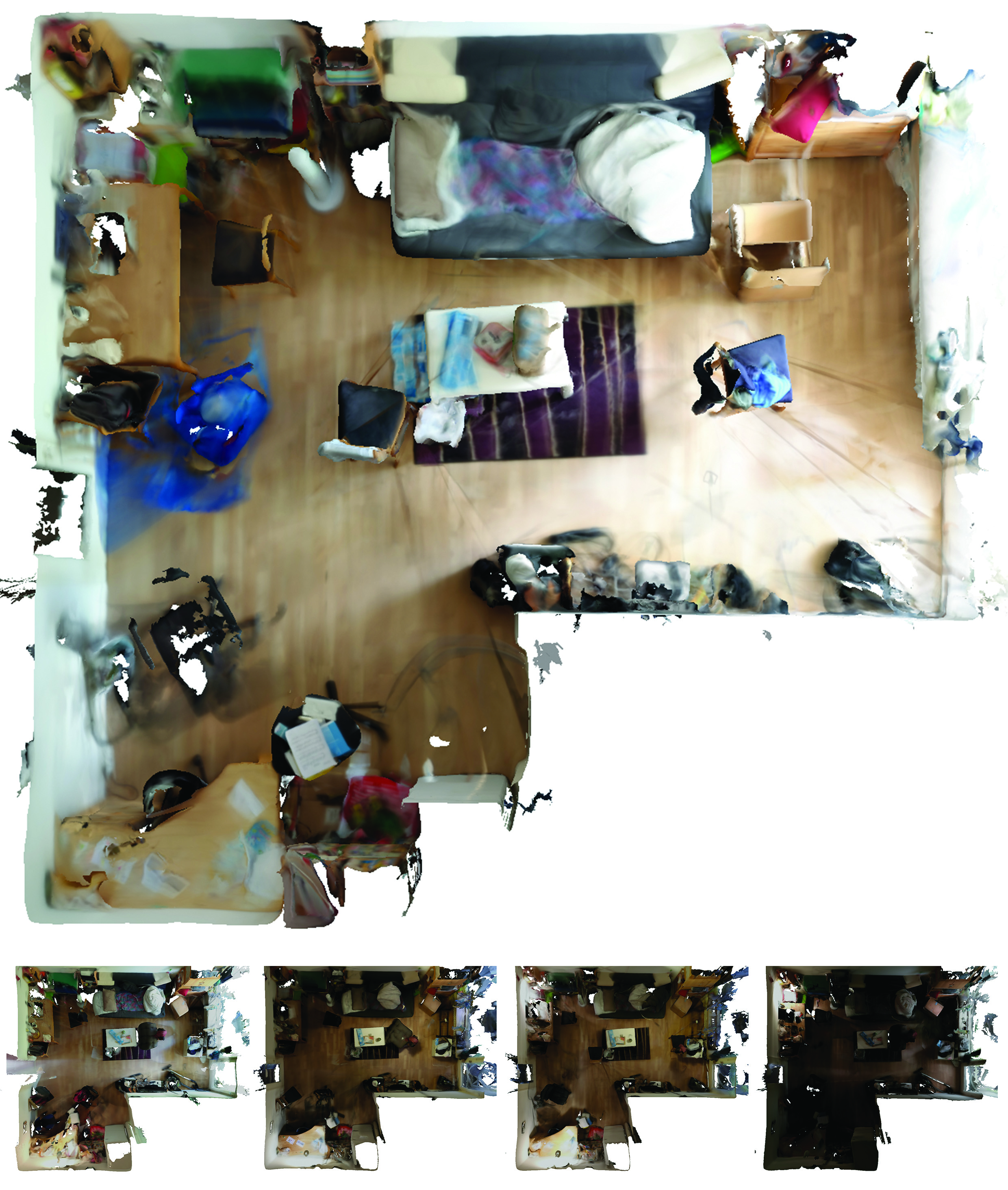} 
    \caption{3D reconstructions of scene 9 of our benchmark dataset.}
    \label{fig:benchmark_overview_s9} 
\end{figure}

\begin{figure}[!p]
    \centering
    \includegraphics[width=\linewidth]{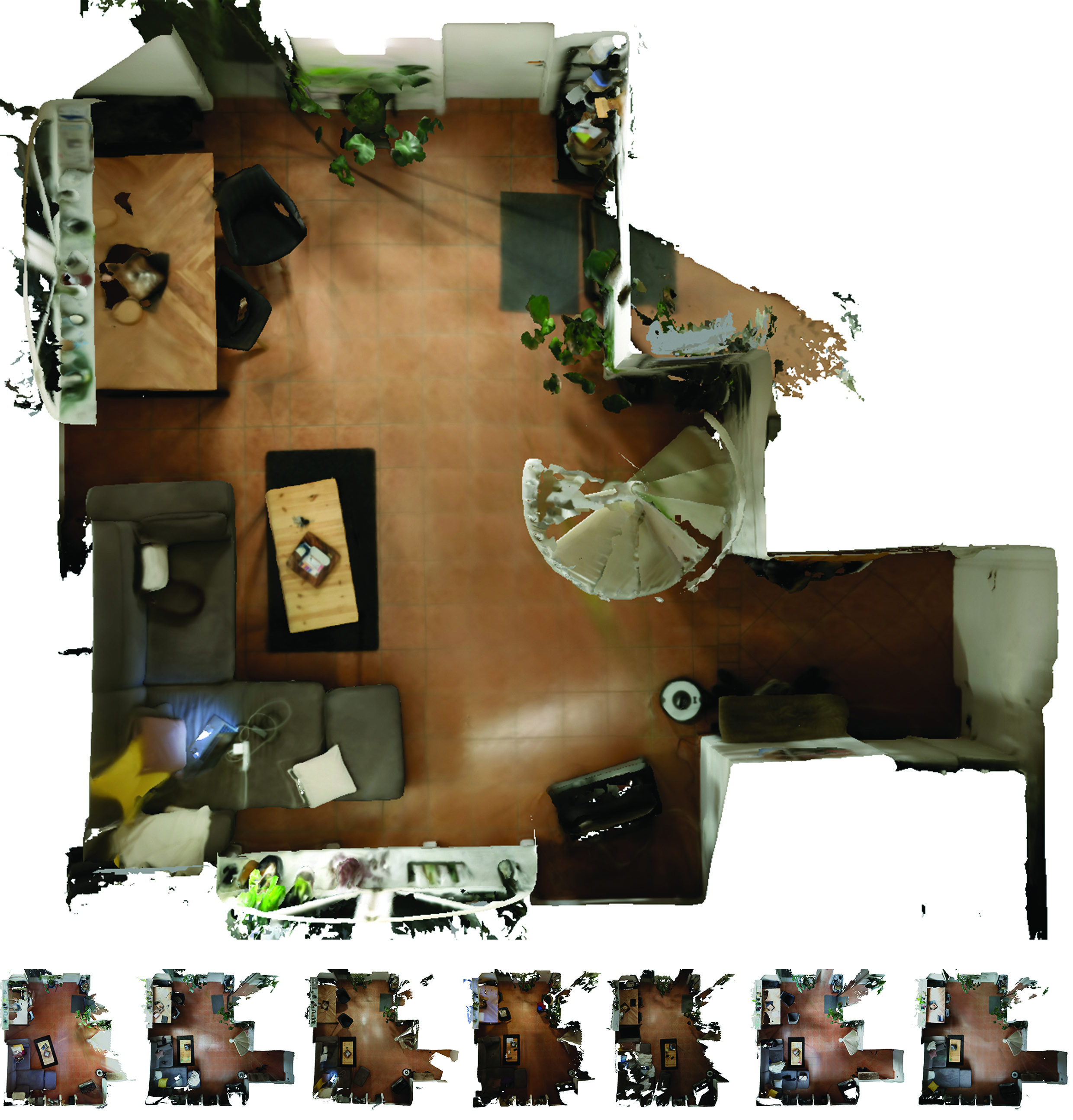} 
    \caption{3D reconstructions of scene 10 of our benchmark dataset.}
    \label{fig:benchmark_overview_s10} 
\end{figure}

\PAR{Details for HF-Net and Image Retrieval} We trained HF-Net for $50k$ iterations on the $52k$ training images of \emph{\nameDataset}. Image retrieval interpolation results (based on the top 20-NN) achieve very similar performance (see Fig.~\ref{fig:plots_image_retrieval} and Table~\ref{table:relocalization_image_retrieval}).

\begin{table}[!t]
\centering
\caption{Comparing the performance of the image retrieval methods NetVLAD and DenseVLAD with and without 20-NN interpolation (\cf\ Table~3 in the main paper).}
\label{table:relocalization_image_retrieval}
\resizebox{\textwidth}{!}{%
{\renewcommand{\arraystretch}{1.3}%
\begin{tabular}{lccccccc}
\toprule
& \multicolumn{4}{c}{\textbf{Inlier}} & \multicolumn{3}{c}{\textbf{Outlier}} \\
& \multicolumn{4}{c}{\raisebox{0.5ex}{\smash{$\overbrace{\makebox[6.2cm]{}}_{}$}}} & \multicolumn{3}{c}{\raisebox{0.5ex}{\smash{$\overbrace{\makebox[4.5cm]{}}_{}$}}} \\
\textbf{Method} & 
$\mathbf{\mathcal{E}_a(0.05m, 5^{\circ})}$ & 
$(\mathbf{\widetilde{\Delta{}t}}, \mathbf{\widetilde{\Delta{}\theta}})$ &
$\;\mathbf{\mathcal{E}_f(0.05)}\;$ &
$\;\mathbf{\mathcal{E}_f(0.15)}\;$ &
$\;$\textbf{N/A}$\;$ & 
$\mathbf{\bar{\mathcal{E}}_a(0.5m, 25^{\circ})}\quad$ & $\mathbf{\bar{\mathcal{E}}_f(0.5)}$\\
\midrule
NetVLAD \cite{Arandjelovic2016} & 0.0002 & $(0.93, 31.44)$  & 0.006 & \multicolumn{1}{p{1cm}|}{$\;$0.1250 } & 0 & 0.798 & 0.452\\
NetVLAD Interpolation \cite{Arandjelovic2016} & 0.0003 & $(0.88, 38.36)$ & 0.007 & \multicolumn{1}{p{1cm}|}{$\;$0.0999 } & 0 & 0.840 & 0.531\\
DenseVLAD \cite{Torii2015} & 0.0003 & $(0.98, 32.26)$ & 0.008 & \multicolumn{1}{p{1cm}|}{$\;$0.1240 } & 0.006 & 0.772 & 0.520 \\
DenseVLAD Interpolation \cite{Torii2015} & 0.0002 & $(1.00, 50.26)$ & 0.008 & \multicolumn{1}{p{1cm}|}{$\;$0.0967 }  & 0.006 & 0.827 & 0.612\\
\bottomrule
\end{tabular}
}}
\end{table}

\PAR{Implementation Details for Sequence-based Re-Localization} The following provides details about the sequence-based re-localization experiments presented in Fig.~9 of the main paper. We use two different approaches, one for RGB-only (Active Search and D2-Net) and one for RGB-D methods (Grove and Grove v2). For both, a sequence is defined as a consecutive set of frames with known \emph{relative} poses. 
For all our experiments, we use relative poses defined by the ground truth absolute poses. 
Note that this does not provide sequence-based methods with any information about where in the scene the images were taken, but it eliminates the impact of pose tracking errors, \eg\ due to drift in visual odometry or SLAM, from the localization process. 
As such, the experiments presented in the paper represent an upper bound on the performance of sequence-based approaches. 
Closing the gap between this upper bound obtained with ``perfect" relative poses and relative poses computed by an existing odometry/SLAM system remains an open research question. 
However, Fig.~9 in the paper shows that considerable gains are possible, which should make practical implementation of sequence-based re-localization an interesting research topic.

For RGB-only methods, we model a sequence of images with known intrinsics and relative poses as a generalized camera~\cite{Pless2003CVPR}, \ie\ as a camera with multiple centers of projection. 
The 2D-3D matches found for each individual image then allow us to estimate the pose of the generalized camera (\ie\ of all images in the sequence simultaneously) by applying a minimal solver for the generalized perspective-$n$-point pose (gP$n$P) problem~\cite{Kukelova2016CVPR,Sweeney2014ECCV,Ventura2014CVPR,Lee2015IJRR} inside a RANSAC~\cite{Fischler81CACM} loop. 
More precisely, we use a gP$n$P+s solver~\cite{Kukelova2016CVPR} that estimates both the pose of the image trajectory and a scale factor, \ie, our approach could account for scale differences between the global 3D model and the trajectory.

For RGB-D methods that process and relocalize each frame independently, we adopt a different approach.
Specifically, for each sequence we want to evaluate, we first transform the relocalized pose for each frame (which denotes the estimated transformation from that frame's camera pose to the \emph{origin of the reference scene}) into a pose expressed relative to the \emph{last} frame in the sequence. This computation is done by combining the frames' \emph{relative} poses with the relocalization output.
We then cluster the transformed relocalized poses (each of which denotes a possible transformation between the \emph{last frame's camera pose in the current scene} and the \emph{origin of the reference scene}) using an iterative approach.
Typically, as a result of the clustering, there will be a single large cluster of poses that are \emph{mutually similar}, and a number of outliers.
We return, as the relocalization result for the sequence, the centroid of the largest cluster (computed, for robustness, via dual-quaternion blending~\cite{Kavan2006} of the corresponding poses).

We will release the code for both of these approaches, thus enabling researchers to more easily work on sequence-based localization. 

\newcommand{\pwdith}{1.7cm} 
\newcommand{\twdith}{1.0cm}
\newcommand{\evalmeasures}{$\mathcal{E}_f(0.05)$}

\begin{table}[!t]
\centering
\caption{Filter setup for evaluation of different challenges on the test / validation images (see Table~\ref{table:performance_filters}). In the following, $\nu$ is the field of view of a frame (as described in Sec.\ 5.1 of the main paper).}
\label{table:filter_setup}
\resizebox{\textwidth}{!}{%
{\renewcommand{\arraystretch}{1.3}%
\begin{tabular}{p{3.0cm}P{1.8cm}P{\pwdith}P{\pwdith}p{\pwdith}P{\pwdith}P{\pwdith}P{\pwdith}P{\pwdith}}
\toprule
\textbf{Filter} & \textbf{\# Images}  & \textbf{$\rho_v$} & \textbf{$\zeta_s$} & \textbf{$\zeta_g$} & \textbf{$\sigma$}  & \textbf{$\nu$} & \textbf{$\eta$} \\
\midrule
(1) no filter & \num{200159} & &  & \\
(2) default filter & \num{161282} & & & & $> 7.2$ & $[0.2, 8]$ & $\leq 650$\\
\midrule
(3) well-textured & \num{84946} & & & & $> 33$ & $[0.2, 8]$ & $\leq 650$\\
(4) texture-less & \num{84704} & & & & $\leq 33$ & $[0.2, 8]$ & $\leq 650$\\
\midrule
(5) high context & \num{63041} & & & & $> 7.2$ & $> 2.4$ & $\leq 650$\\
(6) medium context & \num{62264} & & & & $> 7.2$ & $[0.9, 2.4]$ & $\leq 650$\\
(7) low context & \num{55344} & & & & $> 7.2$ & $\leq 0.9$ & $\leq 650$\\
\midrule
(8) novel poses & \num{20281} & & & & $> 7.2$ & $[0.2, 8]$ & $> 500$\\
(9) not novel poses & \num{36495} & & & & $> 7.2$ & $[0.2, 8]$ & $\leq 150$\\
\midrule
(10) easy changes & \num{5783} & $> 0.8$ & $\leq 0.1$ & $\leq 30$ & $> 7.2$ & $[0.2, 8]$ & $\leq 650$\\
(11) hard changes & \num{13363} & $\leq 0.7$ & $> 0.4$ & $> 30$ & $> 7.2$ & $[0.2, 8]$ & $\leq 650$\\
\bottomrule
\end{tabular}}}
\end{table}

\begin{table}[!t]
\centering
\caption{Evaluation of the different camera re-localization methods with different setups (described in Table~\ref{table:filter_setup}); the reported numbers are \mbox{$\mathcal{E}_f(0.15)$}.}
\label{table:performance_filters}
\resizebox{\textwidth}{!}{%
{\renewcommand{\arraystretch}{1.3}%
\begin{tabular}{p{3.0cm}P{\twdith}P{\twdith}P{\twdith}P{\twdith}P{\twdith}P{\twdith}P{\twdith}P{\twdith}P{\twdith}P{\twdith}P{\twdith}}
\toprule
\textbf{Method} & \textbf{(1)} & \textbf{(2)} & \textbf{(3)} & \textbf{(4)} & \textbf{(5)} & \textbf{(6)} & \textbf{(7)} & \textbf{(8)} & \textbf{(9)} & \textbf{(10)}  & \textbf{(11)} \\
\midrule
Active Search \cite{Sattler2017} & 0.258  & 0.285  & 0.388  & 0.156  & 0.296  & 0.303  & 0.218  & 0.236  & 0.442  & 0.405  & 0.113 \\
Grove \cite{Cavallari2017} & 0.395  & 0.416  & 0.471  & 0.345  & 0.447  & 0.423  & 0.349  & 0.327  & 0.631  & 0.616  & 0.078 \\
\grovetwo~\cite{Cavallari2019PAMI} & 0.487  & 0.509  & 0.570  & 0.430  & 0.559  & 0.514  & 0.425  & 0.413  & 0.715  & 0.714  & 0.112 \\
HF-Net \cite{Sarlin2019coarse} & 0.103  & 0.113  & 0.162  & 0.054  & 0.129  & 0.132  & 0.063  & 0.074  & 0.239  & 0.226  & 0.022 \\
HF-Net Trained \cite{Sarlin2019coarse} & 0.295  & 0.320  & 0.404  & 0.214  & 0.354  & 0.343  & 0.223  & 0.229  & 0.577  & 0.468  & 0.115 \\
D2-Net \cite{Dusmanu2019CVPR} & 0.513  & 0.544  & 0.608  & 0.448  & 0.630  & 0.559  & 0.406  & 0.407  & 0.775  & 0.735  & 0.244 \\
NetVLAD \cite{Arandjelovic2016} & 0.128  & 0.139  & 0.162  & 0.107  & 0.124  & 0.156  & 0.118  & 0.097  & 0.299  & 0.242  & 0.056 \\
DenseVLAD \cite{Torii2015} & 0.126  & 0.136  & 0.160  & 0.102  & 0.135  & 0.153  & 0.105  & 0.110  & 0.307  & 0.237  & 0.049 \\
\bottomrule
\end{tabular}
}}
\end{table}

\clearpage

\clearpage
\bibliographystyle{splncs04}
\bibliography{egbib}

\begin{thebibliography}{100}
\providecommand{\url}[1]{\texttt{#1}}
\providecommand{\urlprefix}{URL }
\providecommand{\doi}[1]{https://doi.org/#1}

\bibitem{Acharya2019}
Acharya, D., Khoshelham, K., Winter, S.: {BIM-PoseNet: Indoor camera
  localisation using a 3D indoor model and deep learning from synthetic
  images}. Journal of Photogrammetry and Remote Sensing  \textbf{150},
  245--258 (2019)

\bibitem{Anoosheh2019ICRA}
{Anoosheh}, A., {Sattler}, T., {Timofte}, R., {Pollefeys}, M., {Gool}, L.V.:
  {Night-to-Day Image Translation for Retrieval-based Localization}. In:
  {International Conference on Robotics and Automation}. IEEE (2019)

\bibitem{Arandjelovic2016}
Arandjelovi\'c, R., Gronat, P., Torii, A., Pajdla, T., Sivic, J.: {NetVLAD}:
  {CNN} architecture for weakly supervised place recognition. In: {Conference
  on Computer Vision and Pattern Recognition}. IEEE (2016)

\bibitem{Ardeshir2014ECCV}
Ardeshir, S., Zamir, A.R., Torroella, A., Shah, M.: {GIS-Assisted Object
  Detection and Geospatial Localization}. In: Fleet, D., Pajdla, T., Schiele,
  B., Tuytelaars, T. (eds.) {European Conference on Computer Vision}. pp.
  602--617. Springer (2014)

\bibitem{Armeni2017}
Armeni*, I., Sax*, A., Zamir, A.R., Savarese, S.: {Joint 2D-3D-Semantic Data
  for Indoor Scene Understanding}. arXiv:1702.01105  (2017)

\bibitem{Atanasov2016IJRR}
Atanasov, N., Zhu, M., Daniilidis, K., Pappas, G.J.: {Localization from
  semantic observations via the matrix permanent}. {International Journal of
  Robotics Research}  (2016)

\bibitem{Badino_IV11}
Badino, H., Huber, D., Kanade, T.: {Visual Topometric Localization}. In:
  Intelligent Vehicles Symposium (IV) (2011)

\bibitem{Bae2016}
Bae, H., Walker, M., White, J., Pan, Y., Sun, Y., Golparvar-Fard, M.: {Fast and
  scalable structure-from-motion based localization for high-precision mobile
  augmented reality systems}. The Journal of Mobile User Experience
  \textbf{5}(1),  1--21 (2016)

\bibitem{SILDa}
Balntas, V., Frost, D., Kouskouridas, R., Barroso-Laguna, A., Talattof, A.,
  Heijnen, H., Mikolajczyk, K.: (silda): Scape imperial localisation dataset.
  \url{https://image-matching-workshop.github.io/challenge/} (2019)

\bibitem{Balntas2018}
Balntas, V., Li, S., Prisacariu, V.: {RelocNet: Continuous Metric Learning
  Relocalisation using Neural Nets}. In: Ferrari, V., Hebert, M., Sminchisescu,
  C., Weiss, Y. (eds.) {European Conference on Computer Vision}. Springer
  (2018)

\bibitem{Benbihi2019ICCV}
Benbihi, A., Geist, M., Pradalier, C.: {ELF: Embedded Localisation of Features
  in Pre-Trained CNN}. In: {International Conference on Computer Vision}. IEEE
  (2019)

\bibitem{Brachmann2017}
Brachmann, E., Krull, A., Nowozin, S., Shotton, J., Michel, F., Gumhold, S.,
  Rother, C.: {DSAC -- Differentiable RANSAC for Camera Localization}. In:
  {Conference on Computer Vision and Pattern Recognition}. IEEE (2017)

\bibitem{Brachmann2016}
Brachmann, E., Michel, F., Krull, A., Yang, M.Y., Gumhold, S., Rother, C.:
  {Uncertainty-Driven 6D Pose Estimation of Objects and Scenes from a Single
  RGB Image}. In: {Conference on Computer Vision and Pattern Recognition}. IEEE
  (2016)

\bibitem{Brachmann2018}
Brachmann, E., Rother, C.: {Learning Less is More -- 6D Camera Localization via
  3D Surface Regression}. In: {Conference on Computer Vision and Pattern
  Recognition}. IEEE (2018)

\bibitem{Brachmann2019esac}
Brachmann, E., Rother, C.: {Expert Sample Consensus Applied to Camera
  Re-Localization}. In: {International Conference on Computer Vision}. IEEE
  (2019)

\bibitem{Brachmann2019neural}
Brachmann, E., Rother, C.: {Neural-Guided RANSAC: Learning Where to Sample
  Model Hypotheses}. In: {International Conference on Computer Vision}. IEEE
  (2019)

\bibitem{Brachmann2020}
Brachmann, E., Rother, C.: {Visual Camera Re-Localization from RGB and RGB-D
  Images Using DSAC}. arXiv:2002.12324  (2020)

\bibitem{Brahmbhatt2018MapNet}
Brahmbhatt, S., Gu, J., Kim, K., Hays, J., Kautz, J.: {Geometry-Aware Learning
  of Maps for Camera Localization}. In: {Conference on Computer Vision and
  Pattern Recognition}. IEEE (2018)

\bibitem{Bresson2017}
Bresson, G., Alsayed, Z., Yu, L., Glaser, S.: {Simultaneous Localization And
  Mapping: A Survey of Current Trends in Autonomous Driving}. {Transactions on
  Intelligent Vehicles}  \textbf{2}(3),  194--220 (2017)

\bibitem{Bui2019}
Bui, M., Baur, C., Navab, N., Ilic, S., Albarqouni, S.: {Adversarial Networks
  for Camera Pose Regression and Refinement}. International Conference on
  Computer Vision Workshops (ICCVW)  (2019)

\bibitem{CarlevarisBianco2016}
Carlevaris-Bianco, N., Ushani, A.K., Eustice, R.M.: {University of Michigan
  North Campus long-term vision and lidar dataset}. {International Journal of
  Robotics Research}  \textbf{35}(9),  1023--1035 (2016)

\bibitem{Castle2008}
Castle, R., Klein, G., Murray, D.W.: {Video-rate Localization in Multiple Maps
  for Wearable Augmented Reality}. In: International Symposium on Wearable
  Computers. pp. 15--22. IEEE (2008)

\bibitem{Cavallari20193DV}
Cavallari*, T., Bertinetto, L., Mukhoti, J., Torr, P., Golodetz*, S.: {Let's
  Take This Online: Adapting Scene Coordinate Regression Network Predictions
  for Online RGB-D Camera Relocalisation}. In: {International Conference on 3D
  Vision}. IEEE, Qu{\'e}bec, Canada (September 2019)

\bibitem{Cavallari2019PAMI}
Cavallari*, T., Golodetz*, S., Lord*, N.A., Valentin*, J., Prisacariu, V.A.,
  Stefano, L.D., Torr, P.H.S.: {Real-Time RGB-D Camera Pose Estimation in Novel
  Scenes using a Relocalisation Cascade}. {Transactions on Pattern Analysis and
  Machine Intelligence}  (2019)

\bibitem{Cavallari2017}
Cavallari, T., Golodetz, S., Lord, N.A., Valentin, J.P.C., di~Stefano, L.,
  Torr, P.H.S.: {On-the-Fly Adaptation of Regression Forests for Online Camera
  Relocalisation}. {Conference on Computer Vision and Pattern Recognition}
  (2017)

\bibitem{Chang2017}
Chang, A., Dai, A., Funkhouser, T., Halber, M., Nie{\ss}ner, M., Savva, M.,
  Song, S., Zeng, A., Zhang, Y.: {Matterport3D: Learning from RGB-D Data in
  Indoor Environments}. In: {International Conference on 3D Vision}. IEEE
  (2017)

\bibitem{Chen2011CVPR}
Chen, D.M., Baatz, G., K\"{o}ser, K., Tsai, S.S., Vedantham, R.,
  Pylv\"{a}n\"{a}inen, T., Roimela, K., Chen, X., Bach, J., Pollefeys, M.,
  Girod, B., Grzeszczuk, R.: {City-Scale Landmark Identification on Mobile
  Devices}. In: {Conference on Computer Vision and Pattern Recognition}. IEEE
  (2011)

\bibitem{Chen2011}
Chen, D.M., Baatz, G., K{\"o}ser, K., Tsai, S.S., Vedantham, R.,
  Pylv{\"a}n{\"a}inen, T., Roimela, K., Chen, X., Bach, J., Pollefeys, M.,
  Girod, B., Grzeszczuk, R.: {City-scale landmark identification on mobile
  devices}. In: {Conference on Computer Vision and Pattern Recognition}. pp.
  737--744. IEEE (2011)

\bibitem{Clark2017}
Clark, R., Wang, S., Markham, A., Trigoni, N., Wen, H.: {VidLoc: A Deep
  Spatio-Temporal Model for 6-DoF Video-Clip Relocalization}. In: {Conference
  on Computer Vision and Pattern Recognition}. pp. 6856--6864. IEEE (2017)

\bibitem{Dai2017}
Dai, A., Chang, A.X., Savva, M., Halber, M., Funkhouser, T., Nießner, M.:
  {ScanNet: Richly-annotated 3D Reconstructions of Indoor Scenes}. In:
  {Conference on Computer Vision and Pattern Recognition}. IEEE (2017)

\bibitem{Deng2016}
Deng, L., Chen, Z., Chen, B., Duan, Y., Zhou, J.: {Incremental image set
  querying based localization}. Neurocomputing  (2016)

\bibitem{DeTone2018}
DeTone, D., Malisiewicz, T., Rabinovich, A.: {SuperPoint: Self-Supervised
  Interest Point Detection and Description}. In: {Conference on Computer Vision
  and Pattern Recognition Workshops}. IEEE (2018)

\bibitem{Duong2018}
Duong, N.D., Kacete, A., Sodalie, C., Richard, P.Y., Royan, J.: {xyzNet:
  Towards Machine Learning Camera Relocalization by Using a Scene Coordinate
  Prediction Network}. In: {International Symposium on Mixed and Augmented
  Reality}. IEEE (2018)

\bibitem{Dusmanu2019CVPR}
Dusmanu, M., Rocco, I., Pajdla, T., Pollefeys, M., Sivic, J., Torii, A.,
  Sattler, T.: {D2-Net: A Trainable CNN for Joint Detection and Description of
  Local Features}. In: {Conference on Computer Vision and Pattern Recognition}.
  IEEE (2019)

\bibitem{Fischler81CACM}
Fischler, M., Bolles, R.: {Random Sampling Consensus: A Paradigm for Model
  Fitting with Application to Image Analysis and Automated Cartography}.
  Commun. ACM  \textbf{24},  381--395 (1981)

\bibitem{GalvezLopez2011}
G{\'a}lvez-L{\'o}pez, D., Tard{\'o}s, J.D.: {Real-Time Loop Detection with Bags
  of Binary Words}. In: {International Conference on Intelligent Robots and
  Systems}. pp. 51--58. IEEE (2011)

\bibitem{Gee2012}
Gee, A.P., Mayol-Cuevas, W.: {6D Relocalisation for RGBD Cameras Using
  Synthetic View Regression}. In: {British Machine Vision Conference}. pp.
  1--11. BMVA (2012)

\bibitem{Germain2019sparsetodense}
Germain, H., Bourmaud, G., Lepetit, V.: {Sparse-To-Dense Hypercolumn Matching
  for Long-Term Visual Localization}. In: {International Conference on 3D
  Vision}. IEEE (2019)

\bibitem{Glocker2013RealtimeRC}
Glocker, B., Izadi, S., Shotton, J., Criminisi, A.: {Real-time RGB-D camera
  relocalization}. In: {International Symposium on Mixed and Augmented
  Reality}. IEEE (2013)

\bibitem{Glocker2015}
Glocker, B., Shotton, J., Criminisi, A., Izadi, S.: {Real-Time RGB-D Camera
  Relocalization via Randomized Ferns for Keyframe Encoding}. {Transactions on
  Visualization and Computer Graphics}  \textbf{21}(5) (2015)

\bibitem{Golodetz2018}
Golodetz*, S., Cavallari*, T., Lord*, N.A., Prisacariu, V.A., Murray, D.W.,
  Torr, P.H.S.: {Collaborative Large-Scale Dense 3D Reconstruction with Online
  Inter-Agent Pose Optimisation}. {Transactions on Visualization and Computer
  Graphics}  \textbf{24}(11),  2895--2905 (November 2018)

\bibitem{Golodetz2015SPDEMO}
Golodetz*, S., Sapienza*, M., Valentin, J.P.C., Vineet, V., Cheng, M.M.,
  Prisacariu, V.A., K{\"a}hler, O., Ren, C.Y., Arnab, A., Hicks, S.L., Murray,
  D.W., Izadi, S., Torr, P.H.S.: {SemanticPaint: Interactive Segmentation and
  Learning of 3D Worlds}. In: ACM SIGGRAPH Emerging Technologies. p.~22 (2015)

\bibitem{GuzmanRivera2014}
Guzman-Rivera, A., Kohli, P., Glocker, B., Shotton, J., Sharp, T., Fitzgibbon,
  A., Izadi, S.: {Multi-Output Learning for Camera Relocalization}. In:
  {Conference on Computer Vision and Pattern Recognition}. pp. 1114--1121. IEEE
  (2014)

\bibitem{He2018}
He, K., Lu, Y., Sclaroff, S.: {Local Descriptors Optimized for Average
  Precision}. In: {Conference on Computer Vision and Pattern Recognition}. pp.
  596--605. IEEE (2018)

\bibitem{Hua2016}
Hua, B.S., Pham, Q.H., Nguyen, D.T., Tran, M.K., Yu, L.F., Yeung, S.K.:
  {SceneNN: a Scene Meshes Dataset with aNNotations}. In: {International
  Conference on 3D Vision}. IEEE (2016)

\bibitem{Irschara09CVPR}
Irschara, A., Zach, C., Frahm, J.M., Bischof, H.: {From Structure-from-Motion
  Point Clouds to Fast Location Recognition}. In: {Conference on Computer
  Vision and Pattern Recognition}. IEEE (2009)

\bibitem{Kacete2017}
Kacete, A., Wentz, T., Royan, J.: {Decision Forest For Efficient and Robust
  Camera Relocalization}. In: {International Symposium on Mixed and Augmented
  Reality}. pp. 20--24. IEEE (2017)

\bibitem{Kavan2006}
Kavan, L., Collins, S., O'Sullivan, C., Zara, J.: {Dual Quaternions for Rigid
  Transformation Blending}. Tech. Rep. TCD-CS-2006-46, Trinity College Dublin
  (2006)

\bibitem{Kendall2016}
Kendall, A., Cipolla, R.: {Modelling Uncertainty in Deep Learning for Camera
  Relocalization}. In: {International Conference on Robotics and Automation}.
  IEEE (2016)

\bibitem{Kendall2017}
Kendall, A., Cipolla, R.: {Geometric Loss Functions for Camera Pose Regression
  with Deep Learning}. In: {Conference on Computer Vision and Pattern
  Recognition}. IEEE (2017)

\bibitem{Kendall2015}
Kendall, A., Grimes, M., Cipolla, R.: {PoseNet: A Convolutional Network for
  Real-Time 6-DOF Camera Relocalization.} In: {International Conference on
  Computer Vision}. pp. 2938--2946. IEEE (2015)

\bibitem{Kukelova2016CVPR}
Kukelova, Z., Heller, J., Fitzgibbon, A.: {Efficient Intersection of Three
  Quadrics and Applications in Computer Vision}. In: {Conference on Computer
  Vision and Pattern Recognition} (2016)

\bibitem{Larsson2019ICCV}
Larsson, M., Stenborg, E., Toft, C., Hammarstrand, L., Sattler, T., Kahl, F.:
  {Fine-Grained Segmentation Networks: Self-Supervised Segmentation for
  Improved Long-Term Visual Localization}. In: {International Conference on
  Computer Vision}. IEEE (2019)

\bibitem{Laskar2017}
Laskar*, Z., Melekhov*, I., Kalia, S., Kannala, J.: {Camera Relocalization by
  Computing Pairwise Relative Poses Using Convolutional Neural Network}. In:
  {International Conference on Computer Vision Workshops}. pp. 929--938. IEEE
  (2017)

\bibitem{Lee2015IJRR}
Lee, G.H., Li, B., Pollefeys, M., Fraundorfer, F.: {Minimal solutions for the
  multi-camera pose estimation problem}. {International Journal of Robotics
  Research}  \textbf{34}(7),  837--848 (2015)

\bibitem{Li2019ICRA}
Li, J., Meger, D., Dudek, G.: {Semantic Mapping for View-Invariant
  Relocalization}. In: {International Conference on Robotics and Automation}.
  IEEE (2019)

\bibitem{Li2019arXiv}
Li, Q., Zhu, J., Cao, R., Sun, K., Garibaldi, J.M., Li, Q., Liu, B., Qiu, G.:
  {Relative Geometry-Aware Siamese Neural Network for 6DOF Camera
  Relocalization}. arXiv:1901.01049v2  (2019)

\bibitem{Li2018BMVC}
Li, W., Saeedi, S., McCormac, J., Clark, R., Tzoumanikas, D., Ye, Q., Huang,
  Y., Tang, R., Leutenegger, S.: {InteriorNet: Mega-scale Multi-sensor
  Photo-realistic Indoor Scenes Dataset}. In: {British Machine Vision
  Conference}. BMVA (2018)

\bibitem{Li2018RSS}
Li, X., Ylioinas, J., Kannala, J.: {Full-Frame Scene Coordinate Regression for
  Image-Based Localization}. In: {Robotics: Science and Systems} (2018)

\bibitem{Li2018ECCV}
Li, X., Ylioinas, J., Verbeek, J., Kannala, J.: {Scene Coordinate Regression
  with Angle-Based Reprojection Loss for Camera Relocalization}. In:
  Leal-Taix{\'e}, L., Roth, S. (eds.) {European Conference on Computer Vision}.
  Springer (2018)

\bibitem{Li-ECCV-2008}
Li, X., Wu, C., Zach, C., Lazebnik, S., Frahm, J.: {Modeling and Recognition of
  Landmark Image Collections Using Iconic Scene Graphs}. In: Forsyth, D., Torr,
  P., Zisserman, A. (eds.) {European Conference on Computer Vision}. pp. I:
  427--440. Springer (2008)

\bibitem{Li-ECCV-2012worldwide}
Li, Y., Snavely, N., Huttenlocher, D., Fua, P.: {Worldwide pose estimation
  using 3d point clouds}. In: Fitzgibbon, A., Lazebnik, S., Perona, P., Sato,
  Y., Schmid, C. (eds.) {European Conference on Computer Vision}. pp. 15--29.
  Springer (2012)

\bibitem{Lim12CVPR}
Lim, H., Sinha, S.N., Cohen, M.F., Uyttendaele, M.: {Real-Time Image-Based
  6-DOF Localization in Large-Scale Environments}. In: {Conference on Computer
  Vision and Pattern Recognition}. IEEE (2012)

\bibitem{Lu2016}
Lu, G., Yan, Y., Kolagunda, A., Kambhamettu, C.: {A Fast 3D Indoor-Localization
  Approach Based on Video Queries}. In: MultiMedia Modeling. pp. 218--230
  (2016)

\bibitem{Lynen2015RSS}
Lynen, S., Sattler, T., Bosse, M., Hesch, J., Pollefeys, M., Siegwart, R.: {Get
  Out of My Lab: Large-scale, Real-Time Visual-Inertial Localization}. In:
  {Robotics: Science and Systems} (2015)

\bibitem{Maddern2017}
Maddern, W., Pascoe, G., Linegar, C., Newman, P.: {1 year, 1000km: The Oxford
  RobotCar Dataset}. {International Journal of Robotics Research}
  \textbf{36}(1),  3--15 (2017)

\bibitem{Massiceti2017}
Massiceti, D., Krull, A., Brachmann, E., Rother, C., Torr, P.H.S.: {Random
  Forests versus Neural Networks -- What's Best for Camera Localization?} In:
  {International Conference on Robotics and Automation}. IEEE (2017)

\bibitem{Melekhov2017}
Melekhov, I., Ylioinas, J., Kannala, J., Rahtu, E.: {Image-based Localization
  using Hourglass Networks}. In: {International Conference on Computer Vision
  Workshops}. pp. 879--886. IEEE (2017)

\bibitem{Meng2016}
Meng, L., Chen, J., Tung, F., Little, J.J., de~Silva, C.W.: {Exploiting Random
  RGB and Sparse Features for Camera Pose Estimation}. In: {British Machine
  Vision Conference}. BMVA (2016)

\bibitem{Meng2017IROS}
Meng, L., Chen, J., Tung, F., Little, J.J., Valentin, J., de~Silva, C.W.:
  {Backtracking Regression Forests for Accurate Camera Relocalization}. In:
  {International Conference on Intelligent Robots and Systems}. IEEE (2017)

\bibitem{Meng2018IROS}
Meng, L., Tung, F., Little, J.J., Valentin, J., de~Silva, C.W.: {Exploiting
  Points and Lines in Regression Forests for RGB-D Camera Relocalization}. In:
  {International Conference on Intelligent Robots and Systems}. IEEE (2018)

\bibitem{Nakashima2019}
Nakashima, R., Seki, A.: {SIR-Net: Scene-Independent End-to-End Trainable
  Visual Relocalizer}. In: {International Conference on 3D Vision}. IEEE (2019)

\bibitem{Paucher2010}
Paucher, R., Turk, M.: {Location-based augmented reality on mobile phones}. In:
  {Conference on Computer Vision and Pattern Recognition Workshops}. pp. 9--16.
  IEEE (2010)

\bibitem{Pless2003CVPR}
Pless, R.: {Using Many Cameras as One}. In: {Conference on Computer Vision and
  Pattern Recognition} (2003)

\bibitem{Porav2018ICRA}
{Porav}, H., {Maddern}, W., {Newman}, P.: {Adversarial Training for Adverse
  Conditions: Robust Metric Localisation Using Appearance Transfer}. In:
  {International Conference on Robotics and Automation}. IEEE (2018)

\bibitem{Radwan2018}
Radwan, N., Valada, A., Burgard, W.: {VLocNet++: Deep Multitask Learning for
  Semantic Visual Localization and Odometry}. {Robotics and Automation Letters}
   \textbf{3}(4),  4407--4414 (2018)

\bibitem{Rodas2015}
Rodas, N.L., Barrera, F., Padoy, N.: {Marker-less AR in the Hybrid Room using
  Equipment Detection for Camera Relocalization}. In: Navab, N., Hornegger, J.,
  Wells, W.M., Frangi, A. (eds.) {International Conference on Medical Image
  Computing and Computer-Assisted Intervention}. pp. 463--470. Springer (2015)

\bibitem{Saeedi2019}
{Saeedi}, S., {Carvalho}, E.D.C., {Li}, W., {Tzoumanikas}, D., {Leutenegger},
  S., {Kelly}, P.H.J., {Davison}, A.J.: {Characterizing Visual Localization and
  Mapping Datasets}. In: {International Conference on Robotics and Automation}.
  IEEE (2019)

\bibitem{SalasMoreno2013CVPR}
Salas{-}Moreno, R.F., Newcombe, R.A., Strasdat, H., Kelly, P.H.J., Davison,
  A.J.: {{SLAM++:} Simultaneous Localisation and Mapping at the Level of
  Objects}. In: {Conference on Computer Vision and Pattern Recognition}. IEEE
  (2013)

\bibitem{Sarlin2019coarse}
Sarlin, P.E., Cadena, C., Siegwart, R., Dymczyk, M.: {From Coarse to Fine:
  Robust Hierarchical Localization at Large Scale}. In: {Conference on Computer
  Vision and Pattern Recognition}. IEEE (2019)

\bibitem{Sattler2011}
Sattler, T., Leibe, B., Kobbelt, L.: {Fast Image-Based Localization using
  Direct 2D-to-3D Matching}. In: {International Conference on Computer Vision}.
  pp. 667--674. IEEE (2011)

\bibitem{Sattler2012}
Sattler, T., Leibe, B., Kobbelt, L.: {Improving Image-Based Localization by
  Active Correspondence Search}. In: Fitzgibbon, A., Lazebnik, S., Perona, P.,
  Sato, Y., Schmid, C. (eds.) {European Conference on Computer Vision}. pp.
  752--765. Springer (10 2012)

\bibitem{Sattler2017}
Sattler, T., Leibe, B., Kobbelt, L.: {Efficient \& Effective Prioritized
  Matching for Large-Scale Image-Based Localization}. {Transactions on Pattern
  Analysis and Machine Intelligence}  \textbf{9} (2017)

\bibitem{Sattler2018}
Sattler, T., Maddern, W., Toft, C., Torii, A., Hammarstrand, L., Stenborg, E.,
  Safari, D., Okutomi, M., Pollefeys, M., Sivic, J., Kahl, F., Pajdla, T.:
  {Benchmarking 6DOF Outdoor Visual Localization in Changing Conditions}. In:
  {Conference on Computer Vision and Pattern Recognition}. IEEE, Piscataway, NJ
  (2018)

\bibitem{Sattler2012BMVC}
Sattler, T., Weyand, T., Leibe, B., Kobbelt, L.: {Image Retrieval for
  Image-Based Localization Revisited}. In: {British Machine Vision Conference}.
  BMVA (2012)

\bibitem{Sattler2019}
Sattler, T., Zhou, Q., Pollefeys, M., Leal-Taix{\'e}, L.: {Understanding the
  Limitations of CNN-based Absolute Camera Pose Regression}. In: {Conference on
  Computer Vision and Pattern Recognition}. IEEE (2019)

\bibitem{Schoenberger2018}
Sch{\"o}nberger, J.L., Pollefeys, M., Geiger, A., Sattler, T.: {Semantic Visual
  Localization}. In: {Conference on Computer Vision and Pattern Recognition}.
  IEEE (2018)

\bibitem{Shotton2013}
Shotton, J., Glocker, B., Zach, C., Izadi, S., Criminisi, A., Fitzgibbon, A.:
  {Scene Coordinate Regression Forests for Camera Relocalization in RGB-D
  Images}. In: {Conference on Computer Vision and Pattern Recognition}. pp.
  2930--2937. IEEE (2013)

\bibitem{Silberman2012}
Silberman, N., Hoiem, D., Kohli, P., Fergus, R.: {Indoor Segmentation and
  Support Inference from RGBD Images}. In: Fitzgibbon, A., Lazebnik, S.,
  Perona, P., Sato, Y., Schmid, C. (eds.) {European Conference on Computer
  Vision}. Springer (2012)

\bibitem{Song2015}
Song, S., Lichtenberg, S.P., Xiao, J.: {SUN RGB-D: A RGB-D Scene Understanding
  Benchmark Suite}. In: {Conference on Computer Vision and Pattern
  Recognition}. pp. 567--576. IEEE (2015)

\bibitem{Sweeney2014ECCV}
Sweeney, C., Fragoso, V., H{\"o}llerer, T., Turk, M.: {gDLS: A Scalable
  Solution to the Generalized Pose and Scale Problem}. In: {European Conference
  on Computer Vision} (2014)

\bibitem{Taira2018inloc}
Taira, H., Okutomi, M., Sattler, T., Cimpoi, M., Pollefeys, M., Sivic, J.,
  Pajdla, T., Torii, A.: {InLoc}: Indoor visual localization with dense
  matching and view synthesis. In: {Conference on Computer Vision and Pattern
  Recognition}. IEEE (2018)

\bibitem{Taira2019}
Taira, H., Rocco, I., Sedlar, J., Okutomi, M., Sivic, J., Pajdla, T., Sattler,
  T., Torii, A.: {Is This the Right Place? Geometric-Semantic Pose Verification
  for Indoor Visual Localization}. In: {International Conference on Computer
  Vision}. IEEE (October 2019)

\bibitem{Toft2017}
Toft, C., Olsson, C., Kahl, F.: {Long-term 3D Localization and Pose from
  Semantic Labellings}. In: {International Conference on Computer Vision}. IEEE
  (2017)

\bibitem{Toft2018ECCV}
Toft, C., Stenborg, E., Hammarstrand, L., Brynte, L., Pollefeys, M., Sattler,
  T., Kahl, F.: {Semantic Match Consistency for Long-Term Visual Localization}.
  In: Ferrari, V., Hebert, M., Sminchisescu, C., Weiss, Y. (eds.) {European
  Conference on Computer Vision}. Springer (2018)

\bibitem{Torii2015}
Torii, A., Arandjelovi\'c, R., Sivic, J., Okutomi, M., Pajdla, T.: {24/7 place
  recognition by view synthesis}. In: {Conference on Computer Vision and
  Pattern Recognition}. IEEE (2015)

\bibitem{Torii2011}
Torii, A., Sivic, J., Pajdla, T.: {Visual localization by linear combination of
  image descriptors}. In: {International Conference on Computer Vision
  Workshops}. IEEE (2011)

\bibitem{Valada2018}
Valada*, A., Radwan*, N., Burgard, W.: {Deep Auxiliary Learning for Visual
  Localization and Odometry}. In: {International Conference on Robotics and
  Automation}. IEEE (2018)

\bibitem{Valentin2016}
Valentin, J., Dai, A., Nie{\ss}ner, M., Kohli, P., Torr, P., Izadi, S., Keskin,
  C.: {Learning to Navigate the Energy Landscape}. In: {International
  Conference on 3D Vision}. pp. 323--332. IEEE (2016)

\bibitem{Valentin2015RF}
Valentin, J., Nie{\ss}ner, M., Shotton, J., Fitzgibbon, A., Izadi, S., Torr,
  P.: {Exploiting Uncertainty in Regression Forests for Accurate Camera
  Relocalization}. In: {Conference on Computer Vision and Pattern Recognition}.
  IEEE (2015)

\bibitem{Valentin2015SP}
Valentin, J., Vineet, V., Cheng, M.M., Kim, D., Shotton, J., Kohli, P.,
  Nie{\ss}ner, M., Criminisi, A., Izadi, S., Torr, P.: {SemanticPaint:
  Interactive 3D Labeling and Learning at your Fingertips}. {Transactions on
  Graphics}  \textbf{34}(5), ~154 (2015)

\bibitem{Ventura2014CVPR}
Ventura, J., Arth, C., Reitmayr, G., Schmalstieg, D.: {A Minimal Solution to
  the Generalized Pose-and-Scale Problem}. In: {Conference on Computer Vision
  and Pattern Recognition} (2014)

\bibitem{Walch2017}
Walch, F., Hazırbaş, C., Leal-Taixé, L., Sattler, T., Hilsenbeck, S.,
  Cremers, D.: {Image-based localization using LSTMs for structured feature
  correlation}. In: {International Conference on Computer Vision}. IEEE (2017)

\bibitem{Wald2019RIO}
Wald, J., Avetisyan, A., Navab, N., Tombari, F., Niessner, M.: {RIO: 3D Object
  Instance Re-Localization in Changing Indoor Environments}. In: {International
  Conference on Computer Vision}. IEEE (2019)

\bibitem{Widya2018CVA}
Widya, A.R., Torii, A., Okutomi, M.: {Structure from Motion Using Dense CNN
  Features With Keypoint Relocalization}. Transactions on Computer Vision and
  Applications  \textbf{10}(1), ~6 (2018)

\bibitem{Wijmans2017}
Wijmans, E., Furukawa, Y.: {Exploiting {2D} Floorplan for Building-scale
  Panorama {RGBD} Alignment}. In: {Conference on Computer Vision and Pattern
  Recognition}. IEEE (2017)

\bibitem{Wu2017}
Wu, J., Ma, L., Hu, X.: {Delving Deeper into Convolutional Neural Networks for
  Camera Relocalization}. In: {International Conference on Robotics and
  Automation}. IEEE (2017)

\bibitem{Yang2019}
Yang*, L., Bai*, Z., Tang, C., Li, H., Furukawa, Y., Tan, P.: {SANet: Scene
  Agnostic Network for Camera Localization}. In: {International Conference on
  Computer Vision}. IEEE (2019)

\bibitem{Zamir2010ECCV}
Zamir, A.R., Shah, M.: {Accurate Image Localization Based on Google Maps Street
  View}. In: Daniilidis, K., Maragos, P., Paragios, N. (eds.) {European
  Conference on Computer Vision}. Springer (2010)

\bibitem{Zhang2006TDPVT}
Zhang, W., Kosecka, J.: {Image based Localization in Urban Environments}. In:
  3DIM-PVT. IEEE (2006)

\bibitem{Zhou2019ARXIV}
Zhou, Q., Sattler, T., Pollefeys, M., Leal-Taixe, L.: {To Learn or Not to
  Learn: Visual Localization from Essential Matrices}. In: {International
  Conference on Robotics and Automation}. IEEE (2020)

\end{thebibliography}

\end{document}